\documentclass{article} %
\usepackage{iclr2027_conference,times}

\usepackage{amsmath,amsfonts,bm}

\def\eqref#1{equation~\ref{#1}}

\usepackage{hyperref}
\hypersetup{
    colorlinks=true,
    linkcolor=red!70!black,
    citecolor=green!50!black,
    urlcolor=cyan!50!black,
    filecolor={[rgb]{0,0.5,0.5}},
    runcolor={[rgb]{0,0.7,0.7}},
    menucolor=red!70!black,
    pdfborder={0 0 0}
}
\usepackage{url}
\usepackage{subcaption}
\usepackage{graphicx} 
\usepackage{booktabs,tabularx,colortbl}
\usepackage{wrapfig}
\usepackage{float}
\usepackage{multirow}

\fancypagestyle{widfirstpage}{%
  \fancyhead[R]{\href{https://github.com/NUS-HPC-AI-Lab/WiD}{\normalfont https://github.com/NUS-HPC-AI-Lab/WiD}}%
}

\title{Beyond the Estimate: Wiener-Informed \\ Decoding for Easier Training of \\ Pixel-Space Flow Models}

\author{Ziqiao Wang \& Yang You \\
NUS HPC-AI Lab\\
\texttt{ziqiaow@u.nus.edu, yangyou@nus.edu.sg} \\
}

 \iclrfinalcopy %
\begin{document}

\maketitle
\thispagestyle{widfirstpage}

\begin{abstract}
Learning velocity fields directly in pixel space is difficult and slow. 
We show that coarse second-order image statistics can ease this task: frozen Wiener estimates improve velocity learning, 
with gains extending beyond the linear predictions they explicitly provide. 
Moreover, we observe similar but complementary gains from frozen bases and pixel decoding.
These observations motivate \textbf{W}iener-\textbf{i}nformed \textbf{D}ecoding (WiD), 
which uses statistical knowledge both for direct prediction 
and to organize decoder computation without modifying native blocks. 
WiD combines a \textit{trainable Wiener prediction path with coefficient interfaces} 
that inject features into the native decoder 
and return its predicted coefficient corrections,
with these components initialized with data statistics. 
All introduced parameters, including the spectrum and coordinate transforms, are trained jointly under the original pixel-space velocity objective.
Across model scales and decoder architectures, WiD accelerates training and improves generation quality with modest parameter overhead. 
It also yields smoother gradient trajectories and remains effective with representation alignment. 
WiD improves class-conditional and text-to-image generation across different resolutions.
On ImageNet $256\times256$, PixelDiT-XL with WiD reaches FID 1.53 after 500 epochs, achieving comparable quality to PixelDiT
with 37.5\% fewer training epochs.
Statistical denoisers contribute beyond the estimate: their computational structure can also ease pixel-space training.
\end{abstract}
\section{Introduction}

Pixel-space flow models offer a simpler generative pipeline, but not an easier learning problem.
By operating directly in pixel space, they remove the need for a separate, pretrained autoencoder.
However, the denoising network must operate on noisy high-dimensional pixels while learning natural image structure, which makes optimization slow and challenging~\citep{jit_Li_2026, Ma2026deco}.

Recent work addresses the challenge in two prominent directions.
(i) Prediction-space approaches seek more learnable target parameterizations, such as clean-image prediction~\citep{jit_Li_2026} and rank-asymmetric velocity prediction~\citep{asymflow}. 
(ii) Architectural approaches distribute contextual modeling and dense pixel prediction across specialized components. For example, DeCo~\citep{Ma2026deco} and PixelDiT~\citep{yu2026pixeldit} pair a contextual backbone with a fine-grained pixel decoder.
These advances make pixel generation more tractable by changing the parameterization or computational organization—and motivate a closer look at the pixel denoising task itself.

However, high dimensionality does not imply an absence of exploitable structure. 
Natural images often possess correlations across space and color, whereas the added Gaussian noise is isotropic.
The second-order statistics yield a closed-form linear denoiser—\textit{the Wiener estimator}~\citep{wiener1949,kailath_linear_2000}.
Although this estimator cannot capture the full complexity of natural images, it helps previous studies~\citep{wang2024the, lukoianov25locality} investigate the learned denoising behavior and locality.
Motivated by these observations, we ask a training question: \textit{is it possible to ease pixel-space velocity learning by explicitly providing this statistical estimate?}

\begin{figure}[tbp]
  \centering
     \begin{subfigure}{0.32\linewidth}
        \includegraphics[width=\linewidth]{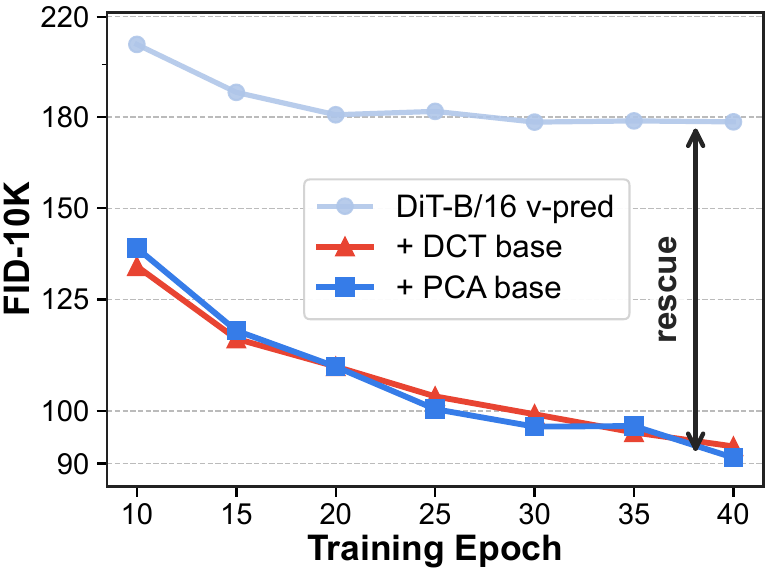}
        \caption{Analytic bases rescue direct velocity prediction DiT in pixel space.}
        \label{fig:fig1a_analytic_prior_rescue}  
     \end{subfigure}
     \hfill
      \begin{subfigure}{0.32\linewidth}
        \includegraphics[width=\linewidth]{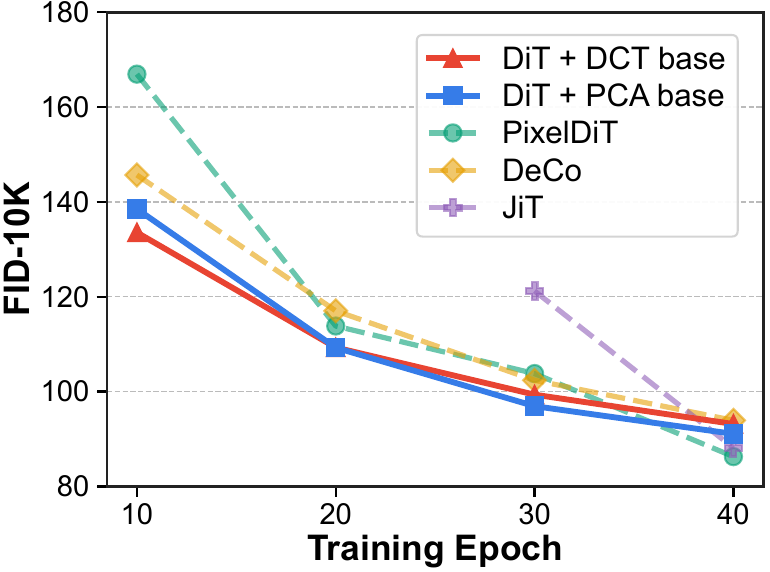}
        \caption{Analytic bases yield competitive quality under native recipes.}
        \label{fig:fig1b_compare}
      \end{subfigure}
      \hfill
      \begin{subfigure}{0.32\linewidth}
        \includegraphics[width=\linewidth]{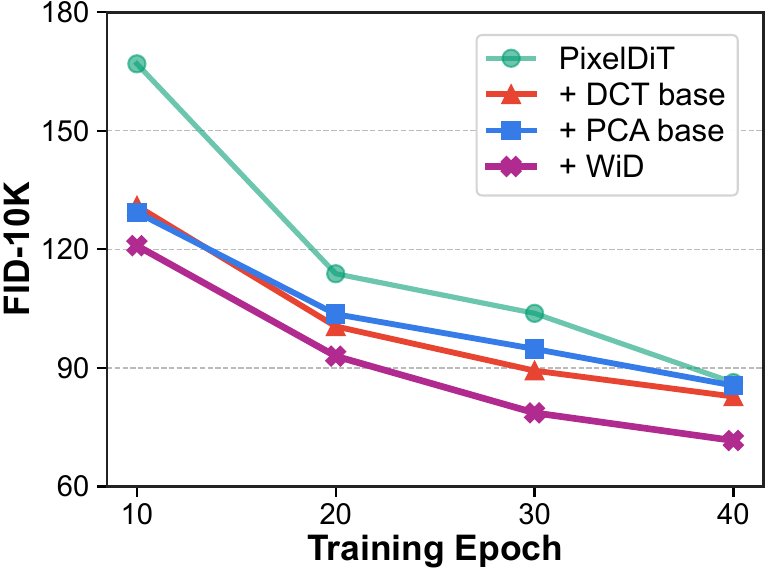}
        \caption{WiD provides stronger and more persistent gains on PixelDiT.}
        \label{fig:fig1c_wid_gains}
      \end{subfigure}
      \caption{
            \textbf{Analytic Wiener bases ease pixel-space velocity learning.}
            We train B-size models for 40 epochs on ImageNet $256{\times}256$ w/o REPA and report FID-10K results w/o CFG.
            \textbf{(a)} With the same DiT-B/16 setup, adding a DCT or PCA Wiener base rescues the velocity prediction baseline.
            \textbf{(b)} DiTs with analytic bases reach the performance range of several pixel-space methods. Each method uses its native training and evaluation recipe.
            \textbf{(c)} When added to native PixelDiT, our Wiener-informed Decoding provides stronger and more sustainable gains than frozen Wiener bases.
            }
      \label{fig:fig1}
\end{figure}

\textbf{Observation:} a fixed estimate already makes a substantial difference. 
We construct a fixed Wiener estimator from statistics to provide an analytic velocity base and train the DiT~\citep{Peebles_2023_dit} to predict the residual correction under the velocity objective.
We consider two typical constructions: global DCT~\citep{DCT1974} for image-wide spectral structure and patch PCA~\citep{ZHANG2010PCA} for local spatial feature.
Both analytic bases rescue the otherwise struggling plain DiT (Figure~\ref{fig:fig1a_analytic_prior_rescue}), bringing generation quality into the range of contemporary approaches (Figure~\ref{fig:fig1b_compare}).
Consequently, even coarse statistical knowledge can ease the burden of velocity learning.

A frozen statistical prediction is not the whole story.
For DiT, our diagnostics further show that much of the reduction in velocity prediction error lies beyond the fitted linear component (Figure~\ref{fig:fig2a_more_than_estimate}).
Thus, the effect of Wiener bases is not confined to correcting a linear prediction.
This leads us to consider the computation behind the estimate: the Wiener estimator transforms a noisy observation into statistical coordinates, adjusts its coefficients according to the noise level, and maps them back to pixels. 
\textit{Can we go a step further and use these operations to organize the denoiser's computation, rather than only adding the resulting estimate to its output?}

Pixel decoding provides a natural and task-aligned interface for this integration.
In PixelDiT and DeCo, the backbone provides semantic context, while the decoder maps noisy pixels to velocity---\textit{the same input--output task} addressed by the Wiener estimator.
We observe similar but complementary gains from frozen Wiener bases and pixel decoding (Figure~\ref{fig:fig2b_decoder_remainder}).
Our input-perturbation tests support the task alignment: the direct pixel pathway shows closer Wiener alignment than the semantic pathway (Figure~\ref{fig:fig2c_pathway_jvp}).
Therefore, we incorporate the estimator's analysis and synthesis operations into the pixel decoder as trainable components, while retaining the semantic backbone architecture.

\textbf{Approach:} from frozen bases to adaptive decoding.
We introduce \textbf{Wiener-informed Decoding (WiD)} that incorporates statistical knowledge in two ways.
(i) \textit{A statistical starting point.}
Using patch PCA statistics, we initialize a direct prediction path to produce the Wiener velocity estimate.
It turns noisy patches into statistical coordinates, 
scales their coefficients according to the noise level,
and maps them back to pixels.
(ii) \textit{Coordinates for decoder computation.}
We also allow noisy-patch coordinates to integrate with the decoder's native input embeddings as additional features.
Alongside its original pixel-space velocity output,
the decoder also predicts a correction to the coefficient-space estimate.
The sum of the Wiener-branch and pixel-head predictions is supervised by the original training objective,
and introduced parameters are trained jointly with the network.

WiD improves training stability and efficiency, and accommodates different pixel-decoder architectures with a modest parameter increase.
Empirically, it reaches comparable generation quality in fewer training steps, with gains that persist further into training than frozen bases
(Figure~\ref{fig:fig1c_wid_gains}).
We validate WiD on class-conditional and text-to-image generation tasks across different resolutions.

\textbf{Contribution Summary.}
We summarize our contributions as follows.
\begin{itemize}
    \item \textbf{Observation.}
    We show that coarse statistical estimates can ease pixel-space velocity learning, with error reductions beyond the fitted linear responses. The observation motivates us to use the computation behind the estimate, rather than only its output.

    \item \textbf{Method.}
    We introduce Wiener-informed Decoding (WiD), which integrates a trainable statistical prediction path with coefficient-based feature processing and output correction.

    \item \textbf{Results.}
    We demonstrate improved training stability, efficiency and generation-quality gains across different pixel-decoder architectures with modest overhead.
\end{itemize}

\section{Preliminaries}
\subsection{Linear Wiener Estimation for Flow Matching}

Consider two independent Gaussian vectors $x\sim\mathcal{N}(\mu_x,C_x)$ and $\epsilon\sim\mathcal{N}(\mu_\epsilon,C_\epsilon)$ in $\mathbb{R}^d$, observed through a linear combination $z=\alpha x+\beta\epsilon$ with known scalar coefficients.
The observation has mean $\mu_z=\alpha\mu_x+\beta\mu_\epsilon$ and covariance $C_z=\alpha^2C_x+\beta^2C_\epsilon$.

Assuming $C_z$ is invertible, their Wiener estimates are
\begin{equation}
\begin{aligned}
    \mathbb{E}[x\mid z] = \mu_x+\alpha C_xC_z^{-1}(z-\mu_z),\qquad \mathbb{E}[\epsilon\mid z]= \mu_\epsilon+\beta C_\epsilon C_z^{-1}(z-\mu_z).
\end{aligned}
\label{eq:wiener_conditional_means}
\end{equation}
Under this Gaussian model, these conditional means minimize the expected squared estimation error and are affine functions of the observed mixture~\citep{kailath_linear_2000}.

For linear flow matching, we use standard Gaussian noise $\epsilon\sim\mathcal{N}(0,I)$ and the interpolation
\begin{equation}
    z_t=tx+(1-t)\epsilon,\qquad v=x-\epsilon,
\label{eq:linear_flow}
\end{equation}
where $t\in[0,1]$ runs from noise to data and $I$ is the identity matrix~\citep{liu2022flowstraightfastlearning, lipman2023flow, Albergo_2025_Stochastic}.
Approximating the data distribution by $x\sim\mathcal{N}(\mu,C)$, define
\begin{equation}
    A_t(C)
    =
    \big[tC-(1-t)I\big]
    \big[t^2C+(1-t)^2I\big]^{-1}.
\label{eq:wiener_velocity_operator}
\end{equation}
Subtracting the two conditional means in~\eqref{eq:wiener_conditional_means} gives the analytic velocity estimate
\begin{equation}
\begin{aligned}
    V_W(z_t,t)
    &:= \mathbb{E}[x-\epsilon\mid z_t,t]\\
    &= \mu+A_t(C)(z_t-t\mu).
\end{aligned}
\label{eq:wiener_velocity_affine}
\end{equation}
This expression is exact under the assumed Gaussian prior; for natural images, it provides a second-order statistical approximation to the conditional velocity.

In practice, we use a linear estimator \textit{without an additive offset}. For this estimator class, the relevant statistic is the uncentered second moment $M=\mathbb{E}[xx^\top]=C+\mu\mu^\top$.
With exact second moments, $A_t(M)z_t$ minimizes the expected squared velocity error among linear estimators without an additive offset~\citep{kailath_linear_2000}.
Using a structured empirical approximation $\widetilde M$,
we obtain
\begin{equation}
    V_W(z_t,t)=A_t(\widetilde M)z_t.
    \label{eq:wiener_velocity_meanfree}
\end{equation}
This formulation requires no explicit mean correction, while retaining the mean's contribution to the second moment.
We explain how to construct such moments and use them in training in Section~\ref{sec:sec3}.

\subsection{Pixel Decoder Interfaces}

Architectures such as PixelDiT and DeCo separate contextual modeling from dense pixel prediction.
Given a conditioning input $y$, their common interface can be written as
\begin{equation}
    s=\mathcal{B}(z_t,t,y),\qquad
    \hat v=\mathcal{D}(z_t,s),
\label{eq:pixel_decoder_interface}
\end{equation}
where $\mathcal{B}$ is a contextual backbone, $s$ denotes the conditioning features, and $\mathcal{D}$ is the pixel decoder.
Although the internal blocks may differ, the interface remains consistent.
The decoder embeds noisy pixels, processes features under contextual modulation, and maps the features to a velocity prediction.
We explain how to incorporate Wiener-informed operations into this interface in Section~\ref{sec:sec4}.

\section{Learning with Analytic Wiener Bases}\label{sec:sec3}
\begin{figure}[htbp]
  \centering
     \begin{subfigure}{0.32\linewidth}
        \includegraphics[width=\linewidth]{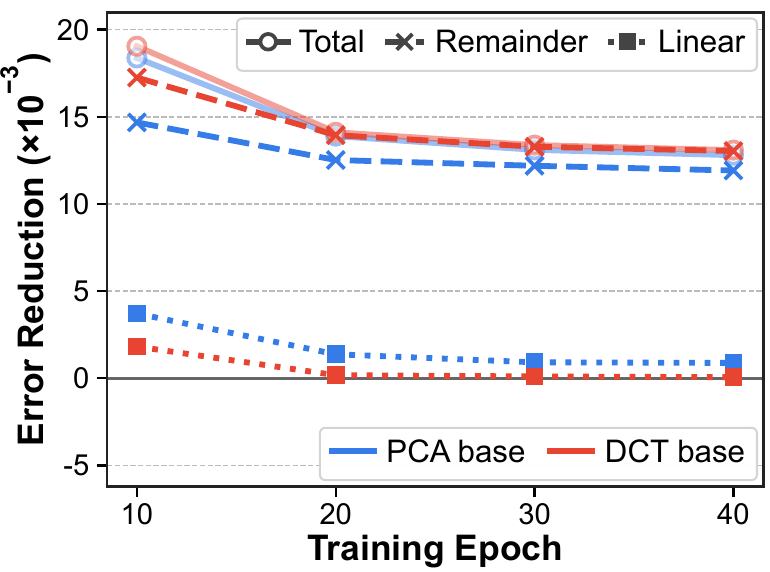}
        \caption{Frozen bases improve beyond the fitted linear responses.}
        \label{fig:fig2a_more_than_estimate}  
     \end{subfigure}
     \hfill
      \begin{subfigure}{0.32\linewidth}
        \includegraphics[width=\linewidth]{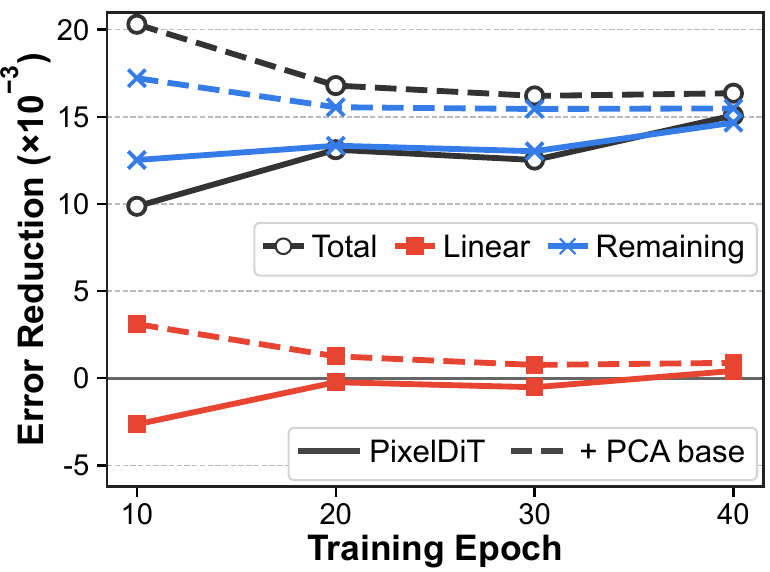}
        \caption{A frozen base provides additional gains with pixel decoding.}
        \label{fig:fig2b_decoder_remainder}
      \end{subfigure}
      \hfill
      \begin{subfigure}{0.32\linewidth}
        \includegraphics[width=\linewidth]{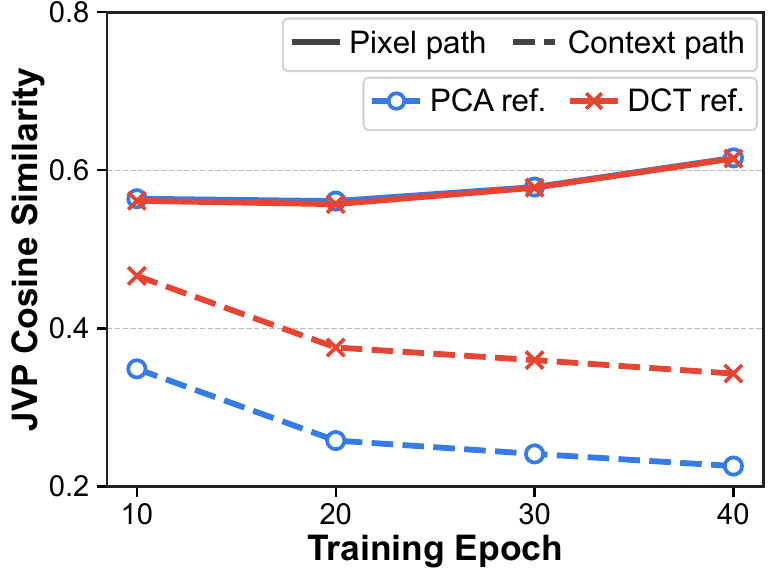}
        \caption{The pixel pathway shows greater alignment to Wiener bases.}
        \label{fig:fig2c_pathway_jvp}
      \end{subfigure}
      \caption{
            \textbf{Prediction gains and pathway alignment.}
            \textbf{(a, b)} Using a well-trained official PixelDiT-XL checkpoint as the prediction reference, we fit affine maps to each model’s predictions and decompose the prediction error into linear and residual components. Relative to plain DiT, we compare how Wiener bases and pixel decoding reduce MSE in each component:
            \textbf{(a)} DiT with frozen PCA or DCT bases;
            \textbf{(b)} PixelDiT and PixelDiT+PCA base evaluated with patch-affine fits.
            \textbf{(c)} PixelDiT JVPs compared with Wiener responses, removing components parallel to perturbation.
            }
      \label{fig:fig2}
\end{figure}
\subsection{Structured Statistical References}
\label{sec:structured_statistical_references}

Let $\widetilde M\in\mathbb R^{d\times d}$ have the eigendecomposition
$\widetilde M=U\Lambda U^\top$, with $U^\top U=I$ and
$\Lambda=\operatorname{diag}(\lambda_1,\ldots,\lambda_d)$,
so that~\eqref{eq:wiener_velocity_meanfree} becomes
\begin{equation}
    V_W(z_t,t)=U\operatorname{diag}\!\big(a_j(t)\big)U^\top z_t,
    \qquad
    a_j(t)=\frac{t\lambda_j-(1-t)}{t^2\lambda_j+(1-t)^2}.
\label{eq:wiener_coordinate_gain}
\end{equation}

\textbf{Estimating the statistics.}
We estimate uncentered second moments from $51{,}200$ randomly sampled
ImageNet training images, using the same RGB center crop to $256\times256$
and $[-1,1]$ scaling as training preprocessing settings without any other augmentation.

\textbf{Two structured bases.}
For $256^2$ RGB images, the full moment matrix has $(3 \times 256^2)^2$
entries. We approximate it in two practical ways.
(i) \emph{Global DCT} retains a $3\times3$ RGB moment at each frequency of the
orthonormal 2-D DCT-II, omitting cross-frequency coupling.
The block eigendecompositions, combined with the DCT, specify $U$ and $\Lambda$.
(ii) \emph{Shared patch PCA} pools all non-overlapping $16^2$ RGB patches
across images and positions, matching the patch size of mainstream pixel-space flow models, and diagonalizes
their shared $768 \times 768$ moment.
The resulting uncentered PCA basis and spectrum are reused for every patch, without modeling cross-patch moments.
Each construction stores $768^2$ moment entries
($256^2\times3^2=768^2$ for DCT), rather than the full image matrix.

\textbf{Residual learning.}
We freeze each base and train the model to predict the residual under the original
velocity objective:
\begin{equation}
\begin{aligned}
    \hat v_\theta(z_t,t,y)&=V_W(z_t,t)+r_\theta(z_t,t,y),\\
    \mathcal L_{\mathrm{FM}}
    &=\mathbb E_{x,y,\epsilon,t}\!\left[
        \frac{1}{d}\left\|\hat v_\theta(z_t,t,y)-(x-\epsilon)\right\|_2^2
       \right],
\end{aligned}
\label{eq:frozen_base_training}
\end{equation}
where $y$ is the conditioning input.
The network still receives the unmodified $z_t$ and no trainable parameters are added. We use the same summed velocity field for sampling and evaluation.

\subsection{Beyond the Additive Linear Prediction}
\label{sec:beyond_additive_prediction}

An analytic base supplies a prediction that the network would otherwise have to learn.
But does it also help the network learn the remaining prediction?
We investigate this by comparing the velocity predictions with a well-trained reference model: the official PixelDiT-XL checkpoint at 320 epoch.

At each fixed $t$, we fit an affine prediction $\ell_f$ from $z_t$ to each model's complete velocity output $f$, and similarly fit $\ell_{\rm ref}$ to the reference $f_{\rm ref}$.
For PCA, the fit is a shared patch-affine map; for DCT, it includes independent RGB-affine maps at each frequency. 
Each affine family contains the corresponding frozen Wiener base as a zero-bias special case, and subtracting the fitted response removes its explicit additive contribution.
We measure errors in decomposed parts:
\begin{equation}
    E_{\rm lin}(f)=\frac{1}{d}\mathbb{E}\|\ell_f-\ell_{\rm ref}\|^2,
    \qquad
    E_{\rm rem}(f)=\frac{1}{d}\mathbb{E}\|(f-\ell_f)-(f_{\rm ref}-\ell_{\rm ref})\|^2.
\label{eq:reference_error_components}
\end{equation}
We fit by least squares on $4{,}096$ validation images and score on $3{,}072$ disjoint images with two fixed noise draws each.
Figure~\ref{fig:fig2a_more_than_estimate} reports the reduction in each error while applying Wiener bases to DiT, averaged over five time points.
We observe that the remaining prediction contributes most of the improvement, and that the linear-error reduction shrinks with training.
These results suggest that a frozen base does more than provide a better starting prediction: 
it may also help the network organize its computation to learn the remaining prediction more effectively.

\subsection{Pixel Decoding and Statistical Prediction}
\label{sec:native_pixel_decoding}

Pixel decoding organizes velocity learning by separating contextual modeling from dense pixel prediction.
We examine its effect using the same reference and patch-affine decomposition as above.
Relative to plain DiT, native PixelDiT's improvement also lies primarily in the remaining prediction (Figure~\ref{fig:fig2b_decoder_remainder}).
However, these gains do not make the statistical estimate redundant.
Adding a frozen PCA base to PixelDiT further reduces both the fitted linear error and the remaining error.
These additional gains generally narrow over training, paralleling the trend in FID gains in Figure~\ref{fig:fig1c_wid_gains}.

We next compare the responses of native PixelDiT's two input pathways with those of the Wiener estimator.
With $t$ and $y$ fixed, we compute each pathway's Jacobian--vector product (JVP) for the same isotropic unit perturbation $\delta$ to $z_t$.
The interface in~\eqref{eq:pixel_decoder_interface} gives
\begin{equation}
    j_{\rm pix}=(\partial_{z_t}\mathcal D)\delta,
    \qquad
    j_{\rm ctx}=(\partial_s\mathcal D)
               (\partial_{z_t}\mathcal B)\delta.
\label{eq:split_pathway_jvp}
\end{equation}
The first holds the contextual features $s$ fixed; the second holds the decoder's direct pixel input fixed.
We compare these output-velocity JVPs with the Wiener responses $A_t(\widetilde M)\delta$, after removing each response's component parallel to $\delta$.
Cosine similarities are averaged over $256$ held-out images and $t\in\{0.1,0.25,0.5,0.75,0.9\}$.
On average, the direct pixel pathway is more aligned with both statistical references than the contextual pathway, and the gap widens over training (Figure~\ref{fig:fig2c_pathway_jvp}).

Together, these observations motivate us to integrate the statistical estimator into pixel decoding.
(i)~Frozen bases and native pixel decoding both improve the remaining prediction, yet their combination yields further gains, showing similar but also complementary benefits.
(ii) The direct pixel pathway's closer alignment with Wiener responses supports the decoder as a natural interface for this integration.
To pursue more persistent gains, we make the estimator's prediction path and coordinate transforms trainable parts of the decoder, allowing them to evolve with learning (Section~\ref{sec:sec4}).

\section{Wiener-informed Decoding}
\label{sec:sec4}

\begin{figure}[htbp]
  \centering
    \includegraphics[width=\linewidth]{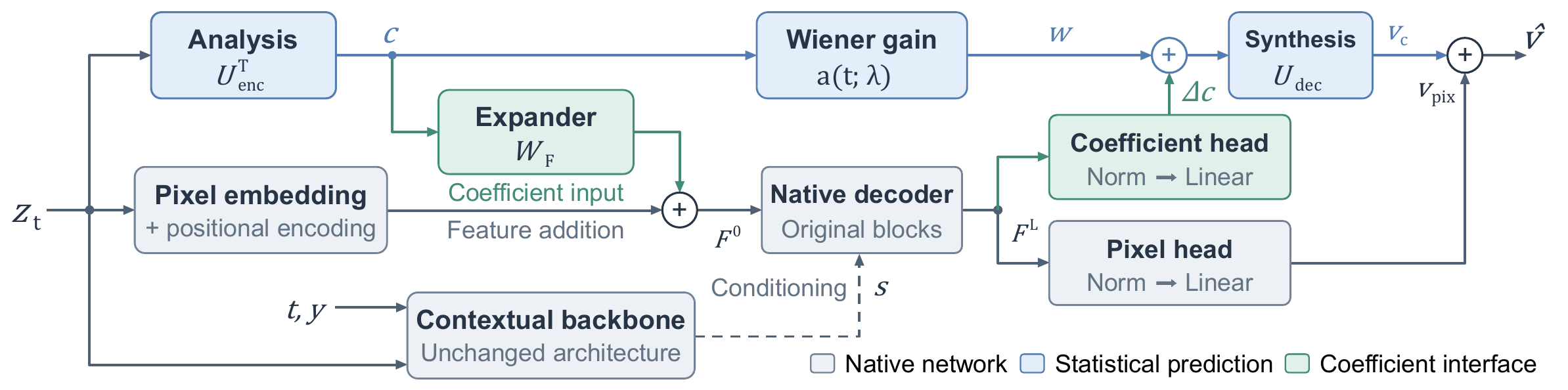}        
    \caption{
            \textbf{Overview of our WiD framework.} 
            WiD integrates statistical operations as trainable components with the native decoder, 
            and allows both branches to interact through coefficient interfaces.
            }
      \label{fig:fig3}
\end{figure}

In this section, we describe how to integrate the statistical estimator into pixel decoding as a trainable component.
As illustrated in Figure~\ref{fig:fig3}, we propose Wiener-informed Decoding (WiD), which couples a trainable Wiener prediction path with coefficient feature processing and output correction.

\subsection{Architecture}
\label{sec:4.1}

\textbf{Wiener path.}
We choose the full shared patch PCA basis for implementation, because it works in the same patch-level spatial space as the pixel decoder.
For a noisy RGB patch $z_{t,i}\in\mathbb R^m$, where $m=3p^2$,
an analysis transform $U_{\rm enc}\in\mathbb R^{m\times m}$ extracts coefficients and applies the Wiener gain:
\begin{equation}
    c_i=U_{\rm enc}^{\top}z_{t,i},
    \qquad
    w_i=a(t;\lambda)\odot c_i.
\label{eq:wid_statistical_coefficients}
\end{equation}
The statistical prediction is obtained by mapping $w_i$ back to pixels through a synthesis transform $U_{\rm dec}\in\mathbb R^{m\times m}$.
Here $U_{\rm enc}$, $U_{\rm dec}$, and $\lambda$ are trainable parameters, and $a(t;\lambda)$ follows \eqref{eq:wiener_coordinate_gain}.

\textbf{Coefficient interaction.} %
The same coefficients, \emph{before} Wiener scaling, also enter the native decoder.
A linear expander $W_F\in\mathbb R^{p^2h\times m}$ maps them to pixel-level features of width $h$, which are added to the native pixel embedding $\mathcal E$:
\begin{equation}
    F_i^0=\mathcal E(z_{t,i})+\operatorname{reshape}(W_Fc_i),
    \qquad F^L=\mathcal T(F^0,s).
\label{eq:wid_feature_interface}
\end{equation}
Here $\mathcal T$ denotes the native decoder blocks operating on features and contextual condition $s$.
The decoder now predicts both a residual pixel-space velocity and a coefficient-space correction $\Delta c$:
\begin{equation}
\begin{aligned}
    \Delta c_i
    &=W_\Delta\operatorname{vec}\!\left(\mathcal N_\Delta(F_i^L)\right)
      +b_\Delta,\\
    \hat v_i
    &=\underbrace{U_{\rm dec}\big[w_i+\Delta c_i\big]}_{\text{Wiener-branch prediction}}
      +\underbrace{\mathcal H_{\rm pix}(F^L)_i}_{\text{native pixel prediction}}.
\end{aligned}
\label{eq:wid_dual_output}
\end{equation}
The normalization $\mathcal N_\Delta$ follows the native output head's granularity.
And the coefficient head maps the normalized features to $m$-dimensional correction vectors with linear projection $W_\Delta$ and bias $b_\Delta$.

Thus, the statistical coordinates participate in both feature processing and velocity prediction, rather than only supplying an additive answer.
We assemble the patch predictions and supervise the sum $\hat v_\theta$ of both predictions with the original velocity objective in~\eqref{eq:frozen_base_training}.
Auxiliary objectives such as REPA~\citep{yu2025representation}, when used, remain those of the underlying training recipe.

\subsection{Initialization}
\label{sec:4.2}

Let $U_0$ and $\lambda_0$ denote the shared patch PCA basis and spectrum estimated in Section~\ref{sec:structured_statistical_references}.
We initialize $U_{\rm enc}$ and $U_{\rm dec}$ with $U_0$, and parameterize the spectrum in log space, initialized to $\log\lambda_0$.
The coefficient correction and native pixel-output projections are both zero-initialized.
Therefore, the initial total prediction $\hat v_\theta$ equals the patch PCA Wiener estimate $V_W$ as a starting point.

At the input of native decoder, we split the content between the two routes. 
Let $E_0$ be the original per-pixel RGB projection and $\mathcal L(E_0)$ its patchwise lift, mapping a patch to flattened pixel features. 
We initialize the pixel projector $E=0.5E_0$ and our expander $W_F=0.5\mathcal L(E_0)U_0$. 
Thus, at initialization, the contributions of two routes are balanced, and the combined embedding satisfies
\begin{equation}
    \mathcal L(E)z_{t,i}
    +W_FU_{enc}^\top z_{t,i}
    =\mathcal L(E_0)z_{t,i}.
\label{eq:wid_initial_feature_identity}
\end{equation}

Biases and positional contributions remain single-copy in the native embedding, so the complete initial features are also preserved.
No special initialization is imposed on the internal decoder blocks.
WiD thus starts from the Wiener prediction while preserving the native decoder's initial input features and block initialization.
All introduced parameters are trained jointly with the network.
The analysis and synthesis transforms evolve independently, without orthogonality or inverse constraints.

\begin{figure}[htbp]
  \centering
     \begin{subfigure}{0.32\linewidth}
        \includegraphics[width=\linewidth]{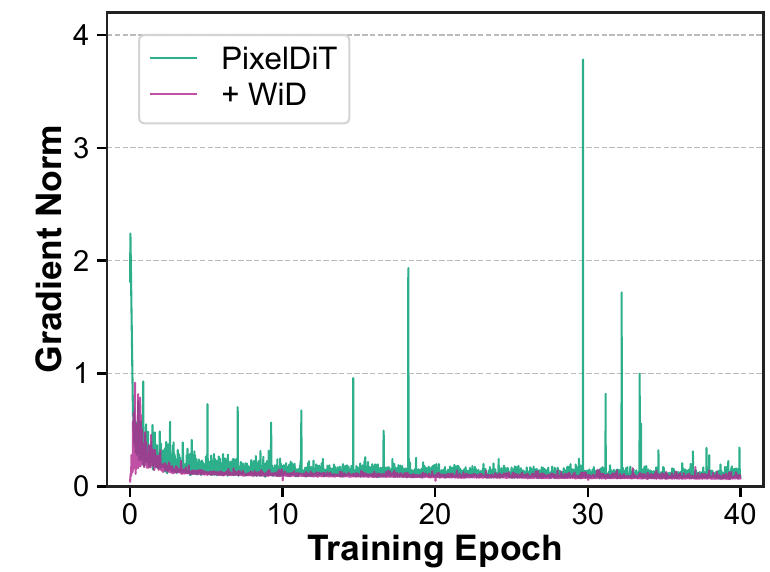}
        \caption{WiD contributes to smoother gradient trajectories on PixelDiT-B.}
        \label{fig:fig4a}  
     \end{subfigure}
     \hfill
      \begin{subfigure}{0.32\linewidth}
        \includegraphics[width=\linewidth]{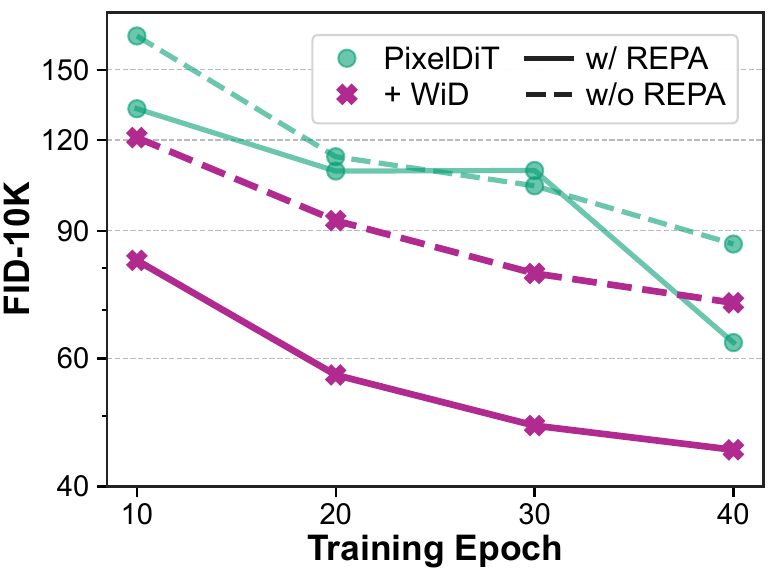}
        \caption{WiD consistently improves generation quality w/ and w/o REPA.}
        \label{fig:fig4b}
      \end{subfigure}
      \hfill
      \begin{subfigure}{0.32\linewidth}
        \includegraphics[width=\linewidth]{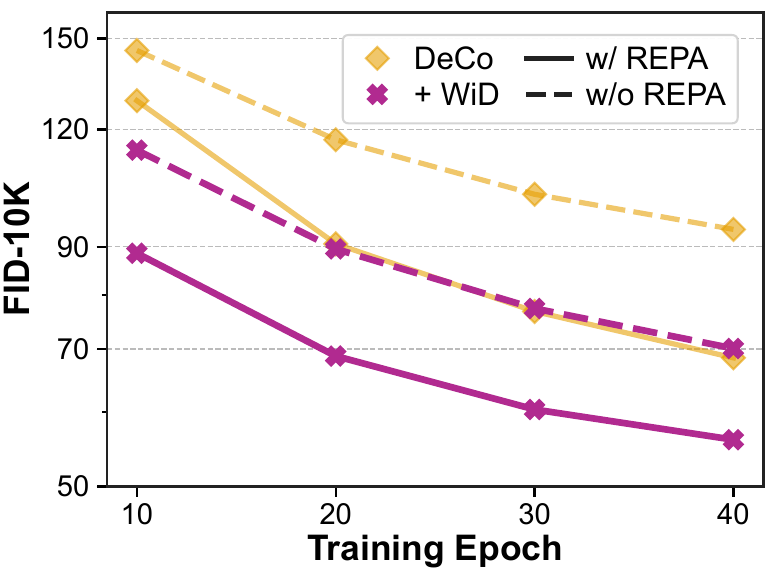}
        \caption{WiD is also effective in improving the generation quality of DeCo.}
        \label{fig:fig4c}
      \end{subfigure}
      \caption{
              \textbf{WiD improves training stability, remains compatible with REPA, and applies across decoder architectures.}
              We compare B-size models trained for 40 epochs on ImageNet $256 {\times} 256$.
              \textbf{(a)} PixelDiT+WiD exhibits smoother pre-clipping gradient norms than native PixelDiT with REPA.
              \textbf{(b,c)} WiD improves both PixelDiT and DeCo and gains persist with REPA.
              }
        \label{fig:fig4}
\end{figure}
\subsection{Training Observations}
\label{sec:4.3}

\textbf{Improved training stability.} 
We report gradient norms before clipping under the same training recipe of PixelDiT-B with and without WiD in Figure~\ref{fig:fig4a}, both using REPA.
Applying WiD leads to smoother gradient trajectories and fewer large excursions than the native model.

\textbf{Compatibility with REPA.}
In the four-way PixelDiT-B comparison in Figure~\ref{fig:fig4b}, 
combining WiD with REPA
yields further gains in generation quality, supporting
their complementary use.

\textbf{Generality across decoders.}
We also test WiD on DeCo-B with a different design of decoder architecture from PixelDiT (Figure~\ref{fig:fig4c}).
Under DeCo's training recipe with its auxiliary frequency-domain objective, WiD consistently improves the generation quality over the native DeCo baseline.

\begin{table}[ht]
  \centering
  \caption{Class-conditional generation on ImageNet $256\times256$. 
  Gray entries denote latent-space models. 
  $\uparrow$ and $\downarrow$ denote higher and lower values are better, respectively.
  Our Wiener-informed Decoding consistently improves the generation quality of PixelDiT-XL during the training process.
  }
  \label{tab:5.1imagenet256}
  \small
  \setlength{\tabcolsep}{3pt}
  \renewcommand{\arraystretch}{1.2}
  \begin{tabularx}{\linewidth}{>{\raggedright\arraybackslash}X*{7}{c}}
    \toprule
    Method & Epochs & \#Params. &
      FID $\downarrow$ & sFID $\downarrow$ & IS $\uparrow$ &
      Precision $\uparrow$ & Recall $\uparrow$ \\
    \midrule
    \textcolor{gray}{REPA-XL~\citep{yu2025representation}}         & \textcolor{gray}{800} & \textcolor{gray}{675M} &
      \textcolor{gray}{1.42} & \textcolor{gray}{4.70} & \textcolor{gray}{305.7} &
      \textcolor{gray}{0.80} & \textcolor{gray}{0.65} \\
    \textcolor{gray}{DDT-XL~\citep{wang2025ddt}}       & \textcolor{gray}{400} & \textcolor{gray}{675M} &
      \textcolor{gray}{1.26} & \textcolor{gray}{--} & \textcolor{gray}{310.6} &
      \textcolor{gray}{0.79} & \textcolor{gray}{0.65} \\
    \textcolor{gray}{RAE-XL~\citep{zheng2026diffusion}}       & \textcolor{gray}{800} & \textcolor{gray}{839M} &
      \textcolor{gray}{1.13} & \textcolor{gray}{--} & \textcolor{gray}{262.6} &
      \textcolor{gray}{0.78} & \textcolor{gray}{0.67} \\
    \arrayrulecolor{gray}\midrule\arrayrulecolor{black}
    ADM~\citep{dhariwal2021diffusion} & 400 & 554M & 4.59 & 5.25 & 186.7 & 0.82 & 0.52 \\
    RIN~\citep{pmlr-v202-jabri23a} & 480 & 410M & 3.42 & -- & 182.0 & -- & -- \\
    VDM++~\citep{kingma2023understanding} & -- & -- & 2.12 & -- & 267.7 & -- & -- \\
    RDM~\citep{teng2024relay} & 400 & \shortstack{553M\,+\,553M} & 1.99 & 3.99 & 260.4 & 0.81 & 0.58 \\
    PixelFlow-XL~\citep{chen2025pixelflowpixelspacegenerativemodels} & 320  & 677M & 1.98 & 5.83 & 282.1 & 0.81 & 0.60 \\
    EPG-XXL/16~\citep{lei2026there} & 600 & 789M & 1.81 & -- & 294.6 & 0.80 & 0.61 \\
    PixNerd-XL~\citep{wang2026pixnerd}     & 320  & 700M & 1.93 & --   & 298.0 & 0.80 & 0.60 \\
    JiT-G~\citep{jit_Li_2026}               & 600  & 2.0B & 1.82 & --   & 292.6 & 0.79 & 0.62 \\
    PixelU-H/16~\citep{guo2026pixelu}      & 600  & 1.17B& 1.63 & 5.04 & 305.9 & 0.79 & 0.64 \\
    DiP-XL/16~\citep{Chen_2026_dip}          & 600  & 631M & 1.79 & 4.59 & 281.9 & 0.80 & 0.63 \\
    FREPix-XL~\citep{lin2026frepixfrequencyheterogeneousflowmatching}         & 320  & 674M & 1.91 & 4.59 & 295.6 & 0.79 & 0.62 \\
    DeCo-XL/16~\citep{Ma2026deco} & 800 & 682M & 1.62 & 4.41 & 301 & 0.80 & 0.62 \\
    \midrule
    PixelDiT-XL/16~\citep{yu2026pixeldit} & 80   & 797M & 2.36 & 5.11 & 282.3 & 0.80 & 0.57 \\
    PixelDiT-XL/16~\citep{yu2026pixeldit} & 320  & 797M & 1.61 & 4.68 & 292.7 & 0.78 & 0.64 \\
    PixelDiT-XL/16~\citep{yu2026pixeldit} & 800  & 797M & 1.54 & 4.49 & 297.0 & 0.78 & 0.65 \\
    PixelDiT-XL/16\textbf{+WiD} & 80  & 805M & 2.09 & 4.73 & 278.7 & 0.79 & 0.60 \\
    PixelDiT-XL/16\textbf{+WiD} & 260 & 805M & 1.61 & 4.67 & 294.2 & 0.77 & 0.65 \\
    PixelDiT-XL/16\textbf{+WiD} & 320 & 805M & 1.58 & 4.87 & 296.1 & 0.77 & 0.65 \\
    PixelDiT-XL/16\textbf{+WiD} & 500 & 805M & \textbf{1.53} & 4.70 & 295.6 & 0.77 & 0.66 \\
    \bottomrule
  \end{tabularx}
\end{table}

\section{Experiments}
\label{sec:sec5}
\subsection{Class-to-Image Generation}
\label{sec:class_to_image}

\textbf{Setup.}
We choose PixelDiT-XL/16\citep{yu2026pixeldit} as the backbone to evaluate our Wiener-informed Decoding on class-conditional ImageNet~\citep{ImageNet} generation.
Following its original training recipe,
we use AdamW optimizer~\citep{kingma2015adam, Loshchilov2017DecoupledWD} with the learning rate of 1e-4 for the first 160 epochs,
and then step down to 1e-5 on ImageNet $256\times256$.
For ImageNet $512\times512$, we fine-tune the 260-epoch $256\times256$ checkpoint for 50 epochs.
The global batch size of 256 and REPA using DINOv2-B~\citep{oquab2024dinov} with weight 0.5 remain the same.

As is stated in Section~\ref{sec:structured_statistical_references}, 
we use 51,200 images (about 4\% of the training set) to compute the spatial patch statistics to initialize WiD.
The computation costs us only \textbf{1 minute} with one GPU and 12 CPU DataLoader workers, which we consider quite practical in large-scale training scenarios.

We generate 50K samples from EMA weights using 100-step FlowDPMSolver~\citep{xie2024sana}
and classifier-free guidance (CFG) ~\citep{ho2022classifierfreediffusionguidance} with interval ~\citep{Kynkaanniemi2024}, 
and report FID~\citep{FID2017}, sFID~\citep{nash2021generating}, IS~\citep{salimans2016improved}, precision, and recall~\citep{Kynkaanniemi2019} using the ADM evaluation
suite~\citep{dhariwal2021diffusion}.

\begin{table}[ht]
  \centering
  \caption{Class-conditional generation on ImageNet $512\times512$. 
  Gray entries denote latent-space models.
  Results demonstrate that WiD is also effective across different resolutions.
  }
  \label{tab:5.1imagenet512}
  \small
  \setlength{\tabcolsep}{3pt}
  \renewcommand{\arraystretch}{1.2}
  \begin{tabularx}{\linewidth}{>{\raggedright\arraybackslash}X*{7}{c}}
    \toprule
    Method & Epochs & \#Params. &
      FID $\downarrow$ & sFID $\downarrow$ & IS $\uparrow$ &
      Precision $\uparrow$ & Recall $\uparrow$ \\
    \midrule
    \textcolor{gray}{REPA-XL~\citep{yu2025representation}}         & \textcolor{gray}{200} & \textcolor{gray}{675M} &
      \textcolor{gray}{2.08} & \textcolor{gray}{4.19} & \textcolor{gray}{274.6} &
      \textcolor{gray}{0.83} & \textcolor{gray}{0.58} \\
    \textcolor{gray}{DDT-XL~\citep{wang2025ddt}}       & \textcolor{gray}{256\,+\,100} & \textcolor{gray}{675M} &
      \textcolor{gray}{1.28} & \textcolor{gray}{4.22} & \textcolor{gray}{305.1} &
      \textcolor{gray}{0.80} & \textcolor{gray}{0.63} \\
    \textcolor{gray}{RAE-XL~\citep{zheng2026diffusion}}       & \textcolor{gray}{400} & \textcolor{gray}{839M} &
      \textcolor{gray}{1.13} & \textcolor{gray}{--} & \textcolor{gray}{259.6} &
      \textcolor{gray}{0.80} & \textcolor{gray}{0.63} \\
    \arrayrulecolor{gray}\midrule\arrayrulecolor{black}
    PixNerd-XL/16~\citep{wang2026pixnerd} & 320\,+\,40 & 700M & 2.45 & 5.19 & 275.8 & 0.79 & 0.63 \\
    JiT-H/32~\citep{jit_Li_2026} & 600 & 956M & 1.94 & -- & 309.1 & -- & -- \\
    PixelU-H/32~\citep{guo2026pixelu} & 600 & 1.17B & 1.92 & 5.98 & 322.1 & 0.80 & 0.58 \\
    DiP-XL/32~\citep{Chen_2026_dip} & -- & 631M & 2.31 & 4.48 & 291.7 & 0.84 & 0.58 \\
    DeCo-XL/16~\citep{Ma2026deco} & 320\,+\,20 & 682M & 2.22 & 4.67 & 290.0 & 0.80 & 0.60 \\
    \midrule
    PixelDiT-XL/16~\citep{yu2026pixeldit} & 320\,+\,40   & 797M & 2.21 & 5.84 & 271.1 & 0.78 & 0.65 \\
    PixelDiT-XL/16~\citep{yu2026pixeldit} & 320\,+\,530  & 797M & 1.81 & 5.61 & 278.6 & 0.78 & 0.67 \\
    PixelDiT-XL/16\textbf{+WiD} & 260\,+\,50 & 805M & \textbf{1.64} & 4.61 & 304.9 & 0.78 & 0.64 \\
    \bottomrule
  \end{tabularx}
\end{table}

\textbf{Main results.}
WiD consistently improves the generation quality of PixelDiT-XL/16 across different training durations.
As shown in Table~\ref{tab:5.1imagenet256}, PixelDiT-XL with WiD reaches
an FID of 2.09 at 80 epochs, outperforming the reported 2.36 of native
PixelDiT at the same duration.
Furthermore, PixelDiT-XL+WiD reaches FID 1.61 at 260 epochs, matching PixelDiT-XL at 320 epochs.
Finally, PixelDiT-XL+WiD achieves FID 1.53 after 500 epochs, saving 37.5\% training epochs of PixelDiT.
We also report the results of generation on ImageNet $512\times512$ in Table~\ref{tab:5.1imagenet512}. 
Our method achieves FID 1.64 after 50 epochs of fine-tuning, outperforming native PixelDiT-XL/16 at 320$+$530 epochs.

The improvement requires only a modest increase in model size (around 1\%) from 797M to 805M parameters.
Together with the B-size comparisons in Section~\ref{sec:4.3},
these results demonstrate that our WiD provides consistent improvement and acceleration across different model scales.

\begin{table}[ht]
  \centering
  \caption{Text-to-image generation at $256\times256$ after 80K training steps.
  We evaluate both methods on 553 GenEval prompts and
  1,065 DPG-Bench prompts, with four images per prompt.
  Higher is better for all scores. WiD improves the performance of DeCo-XXL/16 on both benchmarks.
  }
  \label{tab:5.2text_to_image}
  \small
  \setlength{\tabcolsep}{2pt}
  \renewcommand{\arraystretch}{1.2}
  \begin{tabularx}{\linewidth}{>{\raggedright\arraybackslash}X*{9}{c}}
    \toprule
    \multirow[c]{2}{*}{Method} & \multirow[c]{2}{*}{Steps} &
      \multicolumn{7}{c}{GenEval} & DPG-Bench \\
    \cmidrule(lr){3-9}\cmidrule(lr){10-10}
      & & Sin.Obj. & Two.Obj. & Counting & Colors & Pos. & Color Attr. &
      Overall $\uparrow$ & Average $\uparrow$ \\
    \midrule
    DeCo-XXL/16 & 80K &
      0.85 & 0.32 & 0.27 & 0.74 & 0.12 & 0.24 & 0.43 & 73.1 \\
    DeCo-XXL/16\textbf{+WiD} & 80K &
      {0.92} & {0.45} & {0.33} & {0.81} &
      {0.15} & {0.27} & {0.49} & {78.8} \\
    \bottomrule
  \end{tabularx}
\end{table}

\subsection{Text-to-Image Generation}
\label{sec:text_to_image}

\textbf{Setup.}
We choose DeCo-XXL/16~\citep{Ma2026deco} as the backbone to validate our WiD on Text-to-Image generation task,
since it uses open-source BLIP3-o~\citep{chen2025blip3ofamilyfullyopen} training data.

\textbf{Training recipe.}
Following the original setting of DeCo, we also use the three-stage training strategy.
Firstly, we pretrain the model at $256\times256$ with learning rate 2e-4, global batch size 1536 for 200K steps.
Secondly, we increase the resolution to $512\times512$ and train the model for additional 80K steps with learning rate 1e-4, global batch size 512.
Finally, we fine-tune the model on BLIP3-o-60K at $512\times512$ for 40K steps with learning rate 1e-5, global batch size 512.

\textbf{Statistics computation.}
We also compute statistics over only 51,200 images (around 1\textperthousand~of the training set), including 26,400, 19,200 and 5,600 images from Long-Caption, Short-Caption and JourneyDB datasets.
We use Adams-2nd solver with 25 steps for sampling, and report the performance on GenEval~\citep{ghosh2023geneval} and DPG-Bench~\citep{hu2024ella} benchmarks.

\textbf{Main results.}
Firstly, we report the performance of DeCo-XXL/16 and DeCo-XXL/16+WiD after 80K $256\times256$ pretraining steps in Table~\ref{tab:5.2text_to_image}.
At the early stage of training, WiD consistently improves the performance on both GenEval and DPG-Bench benchmarks at the same training duration.

Furthermore, we report the performance of our DeCo-XXL/16+WiD after full training in Table~\ref{tab:5.2text_to_image_extensive}.
Results show that WiD improves the performance of DeCo-XXL/16 across different resolutions, demonstrating the effectiveness of WiD on large-scale text-to-image generation tasks.
WiD requires only a modest increase of 14M parameters, approximately 1.3\% of the 1.1B-parameter baseline.

\begin{table}[ht]
  \centering
  \caption{Comparison of text-to-image generation at $512\times512$ on GenEval and DPG-Bench.
  $\dagger$~marks results of our test on the public checkpoint of DeCo with its own evaluation protocol.
  WiD also improves the performance of DeCo-XXL across different resolutions.
  }
  \label{tab:5.2text_to_image_extensive}
  \small
  \setlength{\tabcolsep}{1.7pt}
  \renewcommand{\arraystretch}{1.2}
  \begin{tabularx}{\linewidth}{>{\raggedright\arraybackslash}X*{8}{c}}
    \toprule
    \multirow[c]{2}{*}{Method} &
      \multicolumn{7}{c}{GenEval} & DPG-Bench \\
    \cmidrule(lr){2-8}\cmidrule(lr){9-9}
    & Sin.Obj. & Two.Obj. & Counting & Colors & Pos. & Color Attr. &
      Overall $\uparrow$ & Average $\uparrow$ \\
    \midrule
    PixelFlow~\citep{chen2025pixelflowpixelspacegenerativemodels} &
      -- & -- & -- & -- & -- & -- & 0.60 & 77.9 \\
    PixNerd~\citep{wang2026pixnerd} &
      0.97 & 0.86 & 0.44 & 0.83 & 0.71 & 0.53 & 0.73 & 80.9 \\
    PixelDiT-T2I~\citep{yu2026pixeldit} &
      1.00 & 0.94 & 0.70 & 0.90 & 0.53 & 0.65 & 0.78 & 83.7 \\
    \midrule
    DeCo-XXL~\citep{Ma2026deco}  &
      1.00 & 0.92 & 0.72 & 0.91 & 0.80 & 0.79 & 0.86 & 81.4 \\
    DeCo-XXL$^{\dagger}$  &
      0.99 & 0.92 & 0.65 & 0.92 & 0.70 & 0.76 & 0.82 & 82.3 \\
    DeCo-XXL\textbf{+WiD}  &
      {0.98} & {0.93} & {0.75} & {0.94} &
      {0.77} & {0.76} & {0.86} & {82.6} \\
    \bottomrule
  \end{tabularx}
\end{table}

\subsection{Ablations}
\begin{wraptable}{r}{0.45\textwidth}
  \centering
  \caption{
  Ablation experiments on WiD components and initialization on ImageNet $256\times256$.  
  Using PixelDiT-B as the backbone, we train each model for 40 epochs with REPA and report FID, sFID, and IS evaluated on 50K samples from EMA weights without CFG.
  Removing the Wiener prediction, coefficient interface or statistics initialization leads to the reduction of generation quality, 
  demonstrating the effectiveness of components and initialization strategy in our WiD.
  }
  \label{tab:5.3ablation}
  \small
  \setlength{\tabcolsep}{3pt}
  \renewcommand{\arraystretch}{1.2}
  \begin{tabularx}{\linewidth}{@{}>{\raggedright\arraybackslash}X*{3}{c}@{}}
    \toprule
    Method  & FID $\downarrow$ & sFID $\downarrow$ & IS $\uparrow$\\
    \midrule
    PixelDiT-B/16  & 60.01 & 14.10 & 24.89 \\
    PixelDiT-B/16\textbf{+WiD} & 41.71 & 9.18 & 35.14 \\
    \midrule
    w/o Wiener prediction & 49.54 & 10.88 & 29.93 \\
    w/o Statistics init. & 49.02 & 9.85 & 30.35 \\
    w/o Coefficient interface & 50.67 & 9.14 & 28.97 \\
    \bottomrule
  \end{tabularx}
\end{wraptable}

In this section, we conduct ablation experiments of WiD with PixelDiT-B/16 on ImageNet $256\times256$,
to further support claims and analysis in Section~\ref{sec:sec3}\&\ref{sec:sec4}.

\textbf{A statistical starting point counts.} 
We first remove the Wiener prediction,
which serves as a direct pathway of coarse statistical velocity estimation, 
retaining the analysis and synthesis coordinate interactions. 
This leads to the degradation of FID from 41.71 to 49.54, supporting the benefit of the direct statistical prediction pathway.
Secondly, we try random initialization of WiD: initializing $U_{\rm enc}$ and $U_{\rm dec}$ with a random orthogonal matrix, and all spectral values $\lambda$ to 1.
Results in Table~\ref{tab:5.3ablation} also support the benefit of statistical knowledge at initialization.

\textbf{Coordinates for decoder computation matter.} 
Removing the coefficient interfaces between the Wiener and pixel branches increases the FID from 41.71 to 50.67.
Moreover, the ablation that removes Wiener gain still outperforms native PixelDiT-B/16.
These results support our claim that the statistical coordinates are also beneficial for decoder learning.

\section{Related Work}
\label{sec:related_work}

\subsection{Statistical Denoising}
\label{sec:related_statistical_denoising}

\textbf{Statistical priors for image restoration.}
Wiener estimation has a long history in image denoising and restoration~\citep{wiener1949}.
Methods such as BLS-GSM~\citep{Portilla2003} and NL-Bayes~\citep{lebrun2013nonlocal} build on Wiener estimation
using statistics for image denoising.
Similar to statistical operations in our WiD, 
BM3D~\citep{Dabov2007} organizes computation by constructing representations,
applying empirical Wiener filtering to coefficients,
and reconstructing the image through inverse transforms.
Furthermore, Deep Wiener Deconvolution~\citep{dong2020deep_wiener} has proposed to integrate Wiener operations into a
learnable feature-processing pipeline for image deblurring.
Together, these studies show how statistical priors can be constructed and utilized.

\textbf{Statistical structure in diffusion.}
In studies of diffusion models~\citep{sohl2015deep,ho2020denoising,song2021scorebased},
statistical estimates also provide a way to understand learned denoisers.
Linear fits reveal Gaussian structure associated with generalization~\citep{li2024hidden_gaussian}, while Gaussian score
approximations capture moderate and high noise-level behavior and help
accelerate sampling~\citep{wang2024the}.
Data statistics can also explain locality patterns otherwise
attributed to architecture~\citep{lukoianov25locality}.
These findings connect classical estimation to modern generative models,
and inspire us to explore using statistical priors in generative learning.

\textbf{From estimation to generative learning.}
Motivated by these insights, our research takes a step further to study how this statistical knowledge can aid pixel-space velocity learning:
WiD uses the statistics to initialize the prediction path as a structured starting point,
and integrates the computational operations and coordinates into the decoder.

\subsection{Pixel-space Flow Models}
\label{sec:related_pixel_flow}

\textbf{Prediction parameterization.}
Recent approaches ease pixel-space learning by changing what the
network predicts.
JiT~\citep{jit_Li_2026} predicts clean images, while
AsymFlow~\citep{asymflow} combines full-rank data prediction with
low-rank noise prediction to reconstruct the velocity.
Recently, \mbox{$k$-Diff}~\citep{jin2026revisiting} learns a scalar that controls the
data--noise mixture predicted by the network, while retaining
velocity-space supervision.
A simplified linear analysis motivates this parameterization
by relating the preferred mixture to data dimensionality.
Separately, WiD aims to address the similar learning difficulty within velocity prediction.

\textbf{Pixel decoding.}
Another direction is to separate contextual modeling from dense pixel
prediction.
PixNerd~\citep{wang2026pixnerd} uses patch-wise neural fields,
PixelDiT~\citep{yu2026pixeldit} uses pixel transformers, and
DeCo~\citep{Ma2026deco} uses context-conditioned MLP decoders to alleviate the learning difficulty.
WiD builds on this separation: it retains the native backbone and
decoder blocks in PixelDiT and DeCo, adding statistical interfaces
around their existing computation.

\textbf{Connection to preconditioning.}
Preconditioning addresses differences in scale and geometry that can
hinder diffusion and flow training.
EDM~\citep{karras2022elucidating} normalizes network inputs and residual
targets using noise-dependent scaling and a skip connection.
Preconditioned Flow Matching~\citep{ahamed2026preconditioned} instead uses reversible data
transformations to improve the conditioning of intermediate flow
distributions.
WiD is related in the use of statistical structure to ease learning,
but focuses on decoder design: a trainable Wiener prediction path
interacts with the native decoder.

\section{Conclusion}

We present Wiener-informed Decoding (WiD) to improve the velocity learning of pixel-space flow models.
Specifically, we have shown that coarse second-order statistics can ease pixel-space training, 
and the gains from frozen Wiener bases extend beyond their explicit linear predictions.
Motivated by these insights, we developed WiD,
which combines a trainable statistical prediction path with coefficient interfaces that connect it to the native pixel decoder.
We demonstrate that WiD enhances training efficiency and stability while consistently improving generation quality
across model scales and decoder architectures, benefiting both class-conditional and
text-to-image generation with modest parameter overhead. 
These findings suggest that statistical denoisers can contribute beyond the estimate:
the computational organization is also helpful for learning pixel-space generation.

\subsubsection*{Acknowledgments}
We sincerely thank Zhenglin Wan, Pengfei Zhou, Xuanlei Zhao, and Yiqi Zhang from NUS HPC-AI Lab, and Hongzhou Zhu from Tsinghua University for valuable discussions and feedback during this work.
We are grateful to NUSIT's Research Computing group for the computing support with numbers CFP03-CF-088 and CFP05-CF-062.
Yang You's research group is sponsored by ByteDance grant, Alibaba grant, Meta Research Award, AliCloud grant, Apple grant, Google Research Award, Microsoft Research grant, and Adobe gift.
\newpage
\subsection*{AI use statement}

In this work, we used generative AI tools to assist 
providing feedback on research methodology and experiments,
implementing code for methods and figure drawing,
language translation and polishing.
We have not used generative AI tools for 
synthesizing datasets,
developing theoretical models or conceptual frameworks,
formulating mathematical claims,
provide critical ingredients for proving mathematical claims,
writing proofs,
proposing or refining hypotheses,
cleaning and reformating dataset,
supporting qualitative and thematic data analysis,
or interpreting results.
and the rest of the required disclosure tasks are not applicable to this work.
We have reviewed all AI-assisted work. We have verified the accuracy and reliability of code implementation.
We take responsibility for the final content of this work,
including text, claims or artifacts produced with the aid of generative AI.

\subsection*{Reproducibility statement}

Firstly, our experiments are conducted on publicly available datasets, 
and we strictly follow the preprocessing procedures of studies our work is based on. 
Moreover, we provide a comprehensive description of our method (Section~\ref{sec:sec3}\&\ref{sec:sec4}), 
implementation details and training settings (Section~\ref{sec:sec5} and Appendix~\ref{sec:appendix_implementation}).
Finally, we provide detailed evaluation settings and results (Appendix~\ref{sec:appendix_evaluation}).
We will release our training code, model checkpoints, training logs, 
and statistical priors to support reproduction and follow-up research after acceptance.

\bibliography{iclr2027_conference}
\bibliographystyle{iclr2027_conference}
\newpage
\appendix
\section{Additional Experimental Details}
\subsection{Decoder Structures in PixelDiT and DeCo}
\begin{figure}[htbp]
  \centering
     \begin{subfigure}{0.45\linewidth}
        \includegraphics[width=\linewidth]{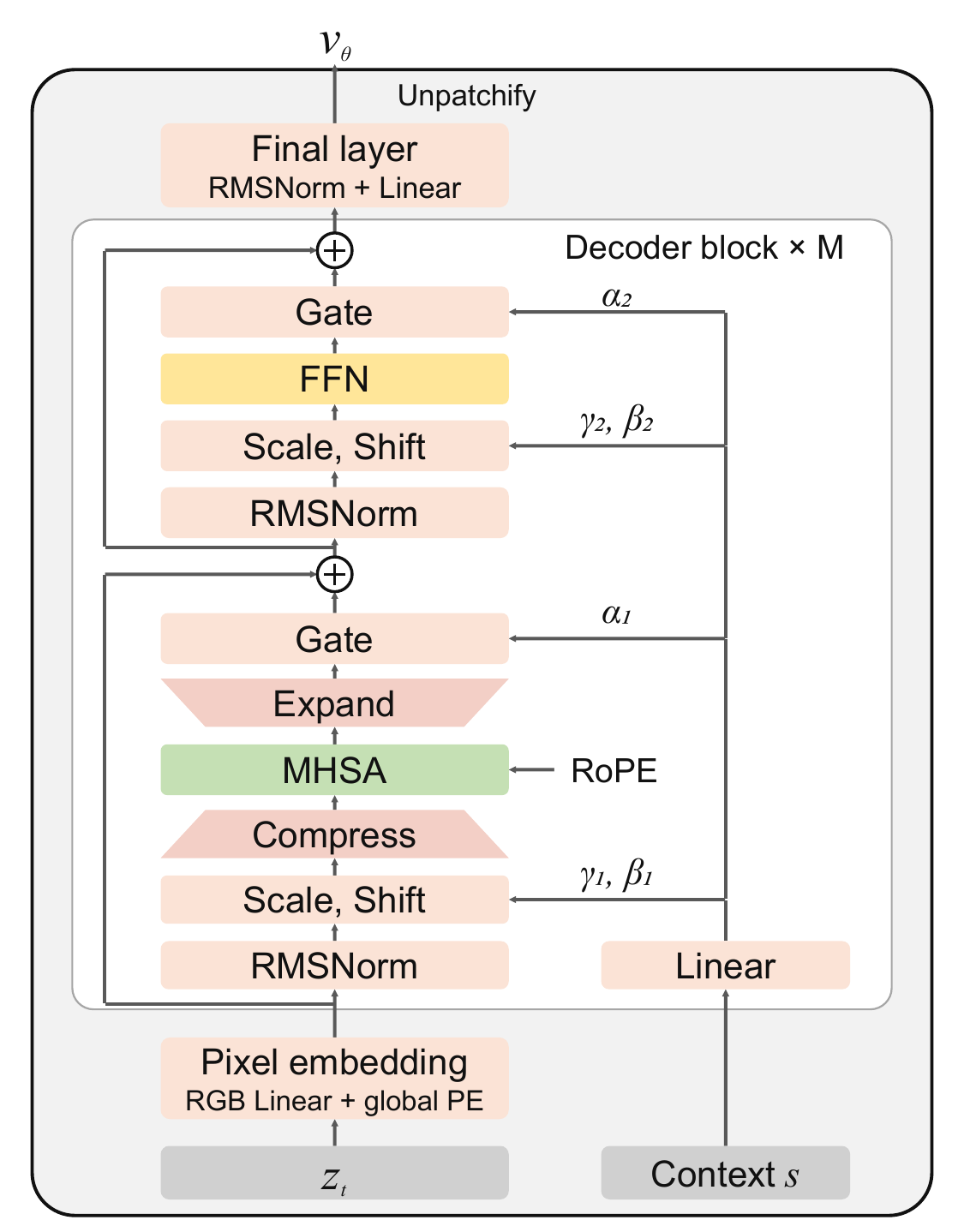}
        \caption{Decoder structure of PixelDiT.}
        \label{fig:app_1a}  
     \end{subfigure}
     \hfill
      \begin{subfigure}{0.45\linewidth}
        \includegraphics[width=\linewidth]{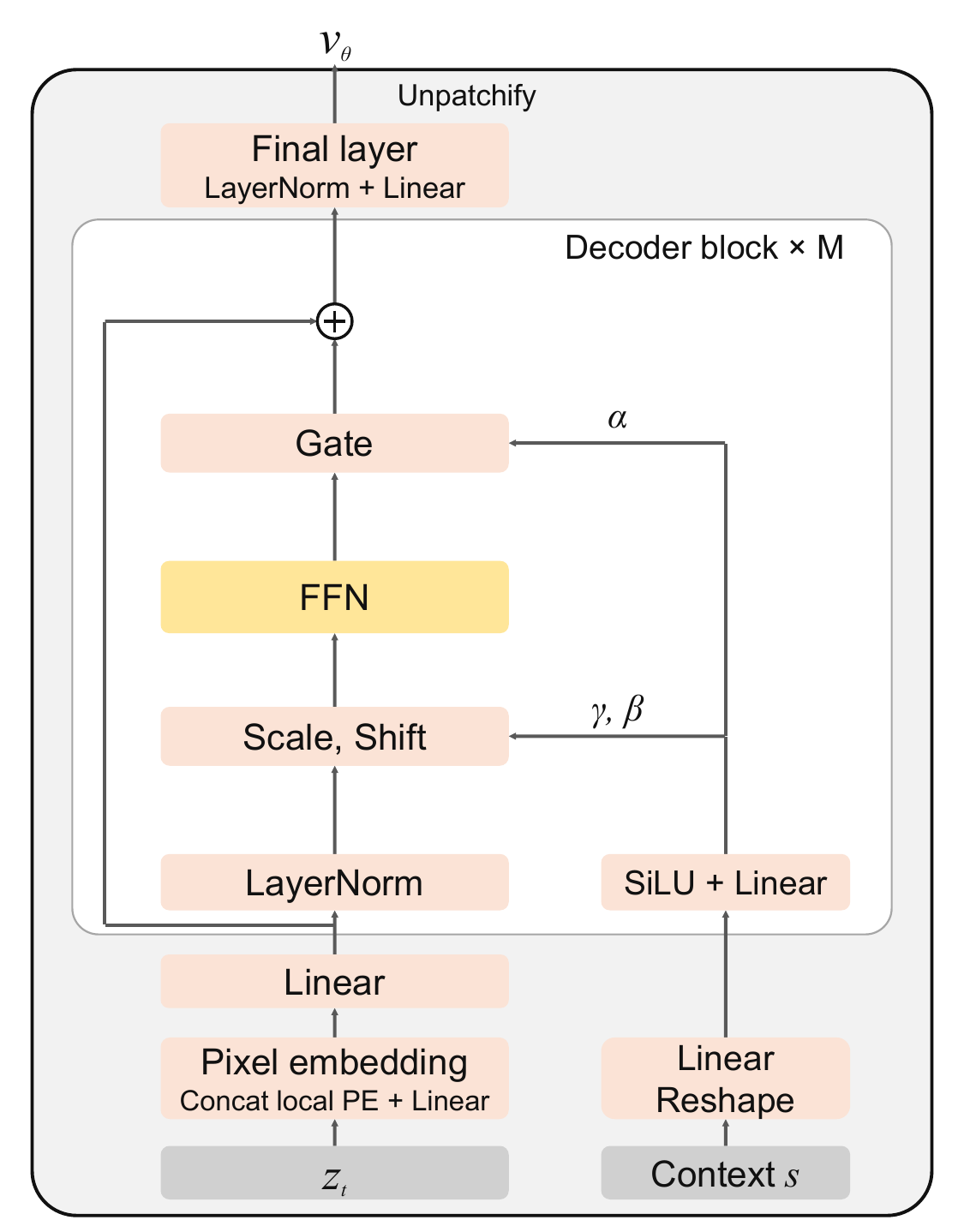}
        \caption{Decoder structure of DeCo.}
        \label{fig:app_1b}
      \end{subfigure}
      \caption{
                Illustration of decoder structures in PixelDiT and DeCo.
              }
        \label{fig:app_1}
\end{figure}

In this work, we use two representative pixel decoders from PixelDiT~\citep{yu2026pixeldit} and DeCo~\citep{Ma2026deco}
to validate our Wiener-informed Decoding approach. 
As shown in Figure~\ref{fig:app_1}, both decoders process the noisy input $z_t$,
use the contextual features $s$ from the backbone to compute scale, shift and gate for pixel-wise modulation,
and finally project normalized features to the velocity~$v_\theta$.
Specifically, PixelDiT decoder involves Multi-Head Self-Attention (MHSA) and Feed-Forward Network (FFN)~\citep{vaswani2017attention} blocks, 
while DeCo decoder mainly uses MLP layers.

\textbf{Input initialization for WiD on DeCo.}
The initialization in Section~\ref{sec:4.2} preserves each decoder's native positional encoding.
PixelDiT adds global positional embeddings after projecting RGB values, whereas DeCo jointly projects RGB values and patch-local cosine positional features.
For pixel $q$ in patch $i$, DeCo's original embedding can be written as
$E_0 z_{t,i,q}+E_{\rm pos}\phi(q)+b$, where $E_0$ contains the RGB columns of the joint projection 
and positional information $\phi(q)$ is shared across patches.
We initialize its RGB projection and the coefficient expander in WiD as
\begin{equation}
    [E_0,\,E_{\rm pos}]\;\longmapsto\;[\tfrac12 E_0,\,E_{\rm pos}],
    \qquad
    W_F=\tfrac12\mathcal L(E_0)U_0,
\label{eq:deco_balanced_init}
\end{equation}
leaving the bias unchanged.
With the full-rank orthogonal basis $U_0$, the injected coefficients recover the other half of the RGB contribution, as in~\eqref{eq:wid_initial_feature_identity}.
The summed features therefore match the original embedding before entering native DeCo decoder blocks at the initial state.

\textbf{Output initialization and normalization for decoders with WiD.}
Both architectures zero-initialize the coefficient-correction and native pixel-output projections, including their biases.
Output normalization follows the native decoder: in DeCo, both heads receive features from its existing per-pixel, non-affine LayerNorm~\citep{ba2016layer}; in PixelDiT, the coefficient head uses per-pixel RMSNorm~\citep{zhang2019root}, matching the native output head's normalization.

\newpage
\subsection{Additional Implementation Details}
\label{sec:appendix_implementation}
\begin{table}[htbp]
  \centering
  \caption{Detailed settings of the five model configurations used in this work.
  Parameter counts refer to the native denoising networks, excluding WiD, REPA projectors and external encoders.
  }
  \label{tab:appendix_model_details}
  \small
  \setlength{\tabcolsep}{2pt}
  \renewcommand{\arraystretch}{1.2}
  \begin{tabularx}{\linewidth}{@{}>{\raggedright\arraybackslash}p{0.18\linewidth}*{5}{>{\centering\arraybackslash}X}@{}}
    \toprule
      & \textbf{DiT-B} & \textbf{PixelDiT-B} & \textbf{PixelDiT-XL} & \textbf{DeCo-B} & \textbf{DeCo-XXL} \\
    \midrule
    \multicolumn{6}{@{}l@{}}{\textit{Architecture}} \\
    DiT depth $N$ & 12 & 12 & 26 & 12 & 16 \\
    Decoder depth $M$ & -- & 2 & 4 & 3 & 3 \\
    Hidden size $D$ & 768 & 768 & 1152 & 768 & 1536 \\
    Attention heads & 12 & 12 & 16 & 12 & 24 \\
    Pixel hidden size & -- & 16 & 16 & 32 & 32 \\
    Patch size $p$ & 16 & 16 & 16 & 16 & 16 \\
    Native params. (M) & 131.4 & 184.7 & 797.4 & 136.0 & 1133.8 \\
    \midrule
    \multicolumn{6}{@{}l@{}}{\textit{Representation alignment}} \\
    Alignment layer & -- & 8 & 8 & 8 & 6 \\
    REPA loss weight & -- & 0.5 & 0.5 & 0.5 & 0.5 \\
    Alignment encoder & -- & DINOv2-B & DINOv2-B & DINOv2-B & DINOv2-B \\
    \midrule
    \multicolumn{6}{@{}l@{}}{\textit{Optimization}} \\
    Training data & ImageNet & ImageNet & ImageNet & ImageNet & BLIP3-o \\
    Global batch size & 256 & 256 & 256 & 256 & 1536$\to$512 \\
    Time sampling & Logit-normal & Logit-normal & Logit-normal & Logit-normal & Logit-normal \\
    Training time shift & 1.0 & 1.0 & 1.0 \& 3.0 & 1.0 & 2.0 \\
    Optimizer & AdamW & AdamW & AdamW & AdamW & AdamW \\
    Adam $(\beta_1,\beta_2)$ & $(0.9,0.999)$ & $(0.9,0.999)$ & $(0.9,0.999)$ & $(0.9,0.999)$ & $(0.9,0.95)$ \\
    EMA decay & 0.9999 & 0.9999 & 0.9999 & 0.9999 & 0.9999 \\
    Condition dropout & 0.1 & 0.1 & 0.1 & 0.2 & 0.1 \\
    Gradient clipping & 1.0 & 1.0 & $1.0\to0.5$ & 1.0 & 1.0 \\
    Lr. ($\times 10^{-4}$) & 1 & 1 & 1$\to$0.1 & 1 & 2$\to$1$\to$0.1 \\
    Weight decay & 0 & 0 & 0 & 0 & 0 \\
    FreqFM loss w. & -- & -- & -- & 1.0 & 1.0 \\
    \midrule
    \multicolumn{6}{@{}l@{}}{\textit{Inference}} \\
    Sampler & FlowDPMSolver & FlowDPMSolver & FlowDPMSolver & Euler & Adams-2nd \\
    Sampling steps & 100 & 100 & 100 & 50 & 25 \\
    Sampling time shift & 1.0 & 1.0 & 1.0 \& 2.0 & 1.0 & 3.0 \\
    \bottomrule
  \end{tabularx}
\end{table}

\textbf{Training settings.} We provide further details of model configurations in Table~\ref{tab:appendix_model_details}. 
When validating frozen Wiener bases and WiD, we strictly follow the original training and inference settings of the native denoising networks.
Particularly, our DiT-B/16 baseline uses PixelDiT-B's 12-block contextual backbone, including RMSNorm, QK normalization~\citep{henry-etal-2020-query}, 2D RoPE~\citep{su2023rope}, and SwiGLU~\citep{shazeer2020swiglu}, rather than the original DiT block design.
We replace the pixel decoder with a single conditioned patch-wise velocity head and use PixelDiT's pixel-space flow-matching recipe.
This construction allows us to compare direct velocity prediction with pixel decoding while keeping the backbone architecture unchanged.
For PixelDiT-XL/16+WiD, we use timeshift 1.0 on ImageNet $256\times256$, and train and sample with 3.0 and 2.0 for ImageNet $512\times512$.

\textbf{Reproducibility.} 
We train PixelDiT-XL/16+WiD on ImageNet $512\times512$, and DeCo-XXL/16+WiD with 8 NVIDIA H200 GPUs.
For PixelDiT-XL/16+WiD on ImageNet $256\times256$, the first 320 epochs are trained on 8 NVIDIA H200 GPUs, and the last 180 epochs are trained on 8 NVIDIA B200 GPUs.
For reproducibility, we fix the random seeds for native-model initialization and for sampling the 51,200 images used to estimate the patch PCA statistics.
We also save each rank's random-number-generator (RNG) state in training checkpoints and restore it when resuming interrupted runs.
We will release our training code, model checkpoints, training logs, and statistical priors to support reproduction and follow-up research.

\newpage
\subsection{Additional Evaluation Details}
\label{sec:appendix_evaluation}

For class-to-image generation on ImageNet-1K 256$\times$256, we strictly follow the native sampling and evaluation protocol of each original backbone.
For comparisons in Figure~\ref{fig:fig1}\&\ref{fig:fig4}, 
we generate 10K images for each method without classifier-free guidance (CFG) and report FID-10K results.
For results of PixelDiT-XL/16+WiD reported in Table~\ref{tab:5.1imagenet256}\&\ref{tab:5.1imagenet512}, we use CFG with guidance interval, 
and we provide more results and detailed settings in Table~\ref{tab:appendix_imagenet256} here. 
At 160 and 200 training epochs, our method also outperforms the native PixelDiT-XL/16 which officially reports FIDs of 1.97 and 1.94. 

\begin{table}[ht]
  \centering
  \caption{We provide more results of PixelDiT-XL/16+WiD with different CFG weights and intervals. 
  }
  \label{tab:appendix_imagenet256}
  \small
  \setlength{\tabcolsep}{3pt}
  \renewcommand{\arraystretch}{1.2}
  \begin{tabularx}{0.95\linewidth}{>{\raggedright\arraybackslash}X*{8}{c}}
    \toprule
    Method & Epochs & CFG & Interval &
      FID $\downarrow$ & sFID $\downarrow$ & IS $\uparrow$ &
      Precision $\uparrow$ & Recall $\uparrow$ \\
    \midrule
    \multicolumn{9}{@{}l@{}}{\textit{ImageNet $\mathit{256}\times\mathit{256}$}} \\
    PixelDiT-XL/16\textbf{+WiD} & 80  & 2.90 & [0.10, 1.00] & 2.09 & 4.73 & 278.7 & 0.79 & 0.60 \\
    PixelDiT-XL/16\textbf{+WiD} & 160 & 2.60 & [0.10, 1.00] & 1.87 & 4.50 & 280.1 & 0.79 & 0.62 \\
    PixelDiT-XL/16\textbf{+WiD} & 200 & 2.75 & [0.10, 0.90] & 1.72 & 4.51 & 295.4 & 0.78 & 0.64 \\
    PixelDiT-XL/16\textbf{+WiD} & 260 & 2.75 & [0.10, 0.90] &1.65 & 4.58 & 296.4 & 0.78 & 0.65 \\
    PixelDiT-XL/16\textbf{+WiD} & 260 & 3.00 & [0.12, 0.90] &1.64 & 4.69 & 295.1 & 0.77 & 0.65 \\
    PixelDiT-XL/16\textbf{+WiD} & 260 & 3.25 & [0.12, 0.70] &1.61 & 4.67 & 294.2 & 0.77 & 0.65 \\
    PixelDiT-XL/16\textbf{+WiD} & 320 & 3.75 & [0.13, 0.55] &1.59 & 4.78 & 295.0 & 0.77 & 0.65 \\
    PixelDiT-XL/16\textbf{+WiD} & 320 & 4.00 & [0.14, 0.55] &1.58 & 4.87 & 296.1 & 0.77 & 0.65 \\
    PixelDiT-XL/16\textbf{+WiD} & 500 & 3.75 & [0.13, 0.55] &1.55 & 4.70 & 295.6 & 0.77 & 0.66 \\
    PixelDiT-XL/16\textbf{+WiD} & 500 & 4.00 & [0.14, 0.55] &1.53 & 4.70 & 295.6 & 0.77 & 0.66 \\
    \midrule
    \multicolumn{9}{@{}l@{}}{\textit{ImageNet $\mathit{512}\times\mathit{512}$}} \\
    PixelDiT-XL/16\textbf{+WiD} & 260\,+\,50 & 3.50 & [0.10, 1.00] & 1.68 & 4.50 & 304.5 & 0.78 & 0.63 \\
    PixelDiT-XL/16\textbf{+WiD} & 260\,+\,50 & 3.50 & [0.10, 0.90] & 1.64 & 4.61 & 304.9 & 0.78 & 0.64 \\
    \bottomrule
  \end{tabularx}
\end{table}

\textbf{Reproducibility.} On ImageNet $256\times256$, we sample 50K images using FlowDPMSolver-100 with the same seed 1600 as the official setting of PixelDiT-XL/16,
and use the same ADM evaluation protocal to compute the scores. 
On ImageNet $512\times512$, we follow the seed 10000 of the official setting of PixelDiT.
Therefore, the results are reproducible with our open-source code and checkpoints. 

\textbf{Computational overhead.} On PixelDiT-XL/16, 
WiD increases the denoiser parameter count from 797.38M to 804.86M and the analytical forward cost from 311.19 to approximately 315.32 GFLOPs, 
corresponding to increases of 0.94\% and 1.33\%, respectively. 
We follow PixelDiT’s linear-layer and attention counting convention 
and additionally account for WiD’s explicit matrix products.

We benchmark denoiser-only inference on a single NVIDIA H200 using EMA weights and BF16, 
with six warm-up iterations followed by 24 synchronized wall-clock measurements. 
At batch size four, WiD reduces forward throughput from 238.8 to 227.6 samples/s for PixelDiT-XL 
and from 260.2 to 245.1 samples/s for our reproduced DeCo-XXL/16 baseline.

With a 25-step Adams-2nd sampler and CFG 4.0, batch-four text-to-image generation throughput is 7.50 images/s for the DeCo-XXL/16 baseline and 7.04 images/s for DeCo-XXL/16+WiD.

\newpage
\section{Additional Diagnostic Details}
\label{sec:appendix_diagnostics}

We expand the five-time averages in Figure~\ref{fig:fig2}.
The analyses below are mainly about frozen Wiener bases and native pixel decoding, before introducing adaptive WiD.

\subsection{Comparison of Prediction Errors}
\label{sec:appendix_time_resolved_errors}

\textbf{Evaluation protocol.}
We use the official PixelDiT-XL epoch-320 EMA checkpoint as the prediction reference and evaluate B-size models at 10, 20, 30 and 40 training epochs.
At each $t\in\{0.1,0.25,0.5,0.75,0.9\}$, we fit on 4,096 ImageNet validation images and score on 3,072 disjoint images, with two fixed noise draws per image.
All comparisons share the same inputs, class labels, and reference predictions, using EMA weights without CFG; the B-size models are trained without REPA.
These diagnostics mainly measure the agreement with a well-trained model.

\begin{table}[ht]
  \centering
  \caption{Time-resolved remaining-error reductions, in units of $10^{-3}$.
  Positive values indicate closer agreement with the reference after removing the fitted affine responses.
  Each block specifies its comparison baseline; all use patch-affine fits except DiT+DCT.}
  \label{tab:appendix_remaining_gains}
  \small
  \setlength{\tabcolsep}{5pt}
  \renewcommand{\arraystretch}{1.2}
  \begin{tabularx}{\linewidth}{@{}>{\raggedright\arraybackslash}X*{5}{>{\centering\arraybackslash}X}@{}}
    \toprule
    Epoch & $t=0.1$ & $t=0.25$ & $t=0.5$ & $t=0.75$ & $t=0.9$ \\
    \midrule
    \multicolumn{6}{@{}l@{}}{\textit{DiT+PCA relative to plain DiT}} \\
    10 & 6.519 & 9.084 & 15.104 & 22.932 & 19.754 \\
    20 & 5.585 & 8.244 & 13.326 & 19.678 & 15.777 \\
    30 & 5.874 & 8.368 & 12.840 & 18.729 & 15.142 \\
    40 & 5.718 & 8.211 & 12.547 & 18.246 & 14.880 \\
    \midrule
    \multicolumn{6}{@{}l@{}}{\textit{DiT+DCT relative to plain DiT}} \\
    10 & 8.955 & 11.067 & 16.279 & 24.144 & 25.817 \\
    20 & 7.653 & 9.980 & 14.353 & 20.131 & 17.621 \\
    30 & 7.043 & 9.389 & 13.563 & 19.176 & 17.251 \\
    40 & 6.945 & 9.264 & 13.267 & 18.694 & 17.030 \\
    \midrule
    \multicolumn{6}{@{}l@{}}{\textit{Native PixelDiT relative to plain DiT}} \\
    10 & 4.719 & 7.812 & 13.172 & 18.363 & 18.547 \\
    20 & 4.929 & 7.612 & 12.648 & 19.150 & 22.380 \\
    30 & 5.455 & 7.957 & 12.423 & 18.706 & 20.593 \\
    40 & 5.653 & 8.266 & 12.862 & 19.951 & 26.588 \\
    \midrule
    \multicolumn{6}{@{}l@{}}{\textit{PixelDiT+PCA relative to native PixelDiT}} \\
    10 & 1.661 & 1.369 & 2.232 & 6.207 & 11.998 \\
    20 & 0.734 & 0.815 & 1.111 & 2.490 & 5.884 \\
    30 & 0.506 & 0.615 & 0.895 & 2.191 & 7.909 \\
    40 & 0.186 & 0.196 & 0.250 & 0.719 & 2.745 \\
    \bottomrule
  \end{tabularx}
\end{table}

\textbf{Fitted responses and remaining predictions.}
At each fixed $t$, we fit an affine map to each model's (including the reference) complete velocity output.
The PCA family is a shared $768\times768$ patch-affine map with bias; the DCT family consists of independent $3\times3$ RGB-affine maps with biases at each of $256\times256$ frequencies.

Each affine family contains its corresponding frozen base, so subtracting the fitted response removes that kind of base's explicit additive contribution.
Note that the remaining prediction can still contain linear dependencies outside the chosen family, such as cross-patch or cross-frequency interactions.

\textbf{Error accounting.}
Let $e_{\rm lin}=\ell_f-\ell_{\rm ref}$ and
$e_{\rm rem}=(f-\ell_f)-(f_{\rm ref}-\ell_{\rm ref})$.
The errors satisfy
\begin{equation}
\begin{aligned}
    E_{\rm tot}(f)
    &=E_{\rm lin}(f)+E_{\rm rem}(f)+E_{\rm cross}(f),\\
    E_{\rm cross}(f)
    &=\frac{2}{d}\mathbb E\langle e_{\rm lin},e_{\rm rem}\rangle.
\end{aligned}
\label{eq:appendix_error_accounting}
\end{equation}
For a baseline $b$ and a comparison model $f$, we report
$\Delta E=E(b)-E(f)$.
Therefore, positive gains over total, linear, or remaining errors indicate smaller errors relative to the reference.

\textbf{Gains beyond the additive prediction.}
Table~\ref{tab:appendix_remaining_gains} reports remaining-error reductions at each time and training epoch.
Both frozen bases and native pixel decoding improve the remaining prediction over plain DiT.
These gains do not require uniformly better fitted linear responses. 
For example, at 40 epochs and $t=0.9$, DiT+DCT and native PixelDiT have higher linear errors but lower remaining and total errors.
All models share the DiT backbone design and global batch size of 256.

\textbf{Additional gains with pixel decoding.}
The last block compares PixelDiT with and without a frozen PCA base.
The base remains useful with a native decoder to 40 epochs: 
all five remaining-error gains are positive.
They shrink overall with training, consistent with the decreasing FID gains in Figure~\ref{fig:fig1c_wid_gains}.

\subsection{Pathway Responses to Input Perturbations}
\label{sec:appendix_pathway_responses}

\textbf{Pathway decomposition.}
We probe native PixelDiT-B using 256 fixed held-out images, one noise per image, EMA weights, and no CFG.
With $t,y$ fixed, automatic differentiation gives the two output-velocity JVPs in~\eqref{eq:split_pathway_jvp}.
The pixel pathway holds the contextual features at their current values; and the contextual pathway holds the decoder's direct noisy input fixed.
The reference here is the analytic response $w=A_t(\widetilde M)\delta$ for either Wiener base.

\textbf{Comparing pathway alignment.}
Following Figure~\ref{fig:fig2c_pathway_jvp}, we remove each response's component parallel to the isotropic unit perturbation $\delta$ before comparison:
\begin{equation}
    j^{\perp}=j-\frac{\langle j,\delta\rangle}{\|\delta\|^2}\delta,
    \qquad
    w^{\perp}=w-\frac{\langle w,\delta\rangle}{\|\delta\|^2}\delta.
\label{eq:appendix_direction_removal}
\end{equation}
We compute cosine similarities per image and report the mean difference between the pixel and contextual pathways.
Positive values favor the pixel pathway alignment with the reference.

\begin{table}[!htbp]
  \centering
  \caption{Time-resolved differences in pathway alignment with Wiener responses.
  Entries are the mean cosine to the reference for the direct pixel pathway minus that for the contextual pathway.}
  \label{tab:appendix_pathway_alignment}
  \small
  \setlength{\tabcolsep}{5pt}
  \renewcommand{\arraystretch}{1.2}
  \begin{tabularx}{\linewidth}{@{}>{\raggedright\arraybackslash}X*{5}{>{\centering\arraybackslash}X}@{}}
    \toprule
    Epoch & $t=0.1$ & $t=0.25$ & $t=0.5$ & $t=0.75$ & $t=0.9$ \\
    \midrule
    \multicolumn{6}{@{}l@{}}{\textit{PCA reference}} \\
    10 & $-0.137$ & 0.200 & 0.405 & 0.316 & 0.292 \\
    20 & $-0.101$ & 0.261 & 0.514 & 0.439 & 0.403 \\
    30 & $-0.054$ & 0.307 & 0.548 & 0.458 & 0.430 \\
    40 & 0.048 & 0.400 & 0.591 & 0.481 & 0.431 \\
    \midrule
    \multicolumn{6}{@{}l@{}}{\textit{DCT reference}} \\
    10 & $-0.351$ & 0.027 & 0.273 & 0.254 & 0.272 \\
    20 & $-0.313$ & 0.091 & 0.376 & 0.373 & 0.380 \\
    30 & $-0.259$ & 0.136 & 0.409 & 0.394 & 0.410 \\
    40 & $-0.150$ & 0.228 & 0.452 & 0.418 & 0.412 \\
    \bottomrule
  \end{tabularx}
\end{table}

\textbf{Dependence on noise level.}
Table~\ref{tab:appendix_pathway_alignment} favors the pixel pathway at all four evaluated times from $0.25$ to $0.9$.
At $t=0.1$, however, the contextual pathway shows generally closer alignment with Wiener bases to DCT at every checkpoint and to PCA at 10--30 epochs.
Averaging over all five times, the pixel pathway has higher direction-removed cosine for each reference.

\section{Limitations and Future Work}
In this work, we mainly focus on easing velocity learning for pixel-space image generation tasks.
Actually, we think the perspective of statistical denoising may also be helpful for other flow generative models.
Therefore, explorations of integrating statistical denoisers into other generative tasks are also attractive directions, 
such as latent space generation, and language, audio, as well as video generation.
In the future, we are planning to improve upon our current approach, and take a step further 
to explore the potential of integrating Wiener-informed structures directly into diffusion transformers,
and try to develop next foundation architecture for flow-matching tasks.

\newpage
\section{Additional Visualizations}
\begin{figure}[H]
    \centering
    \includegraphics[width=1\linewidth]{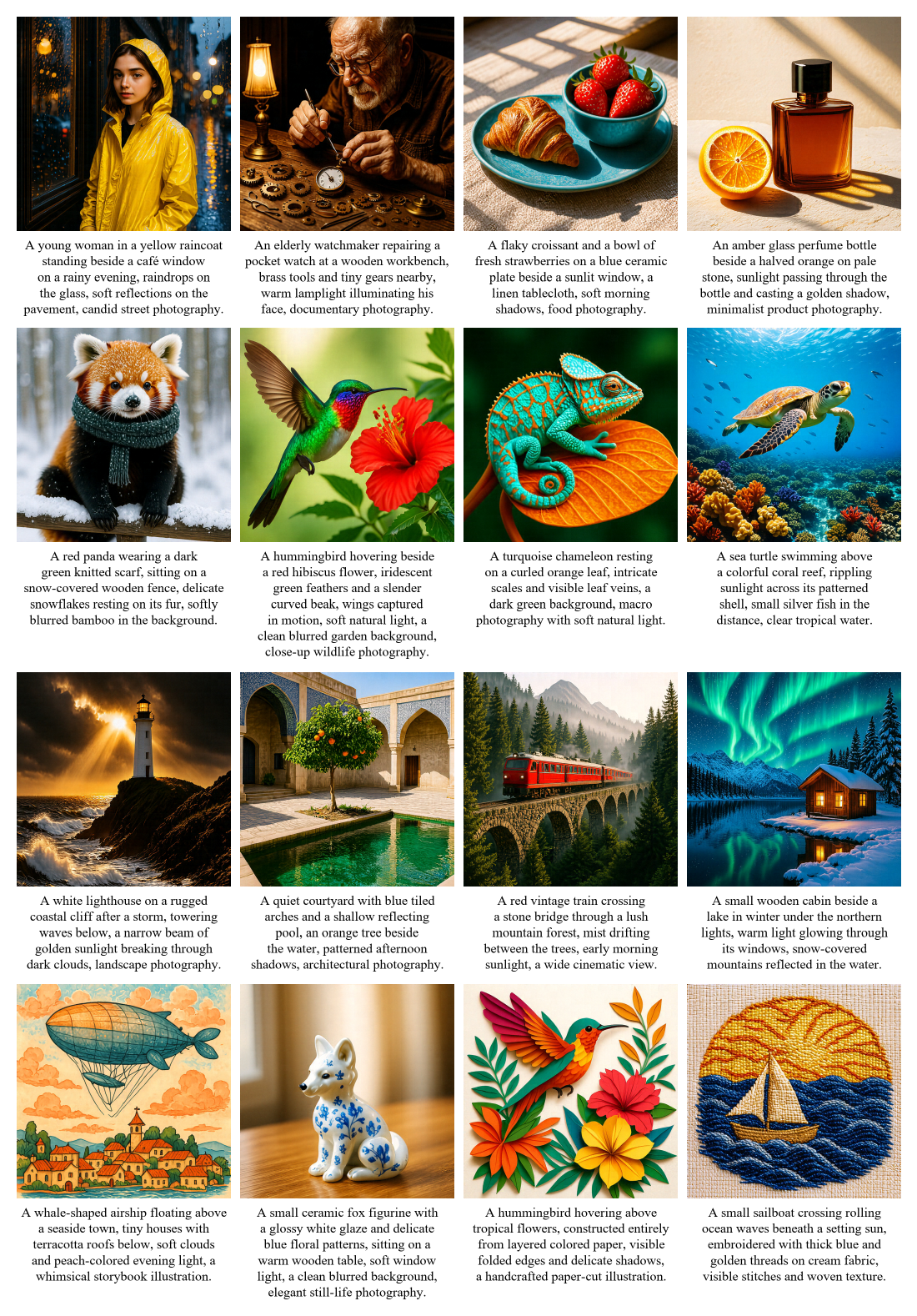}
    \caption{Qualitative results of DeCo-XXL/16+WiD on $512\times512$ text-to-image generation.}
    \label{fig:t2i_samples}
\end{figure}

\begin{figure}[H]
    \centering
    \includegraphics[width=1\linewidth]{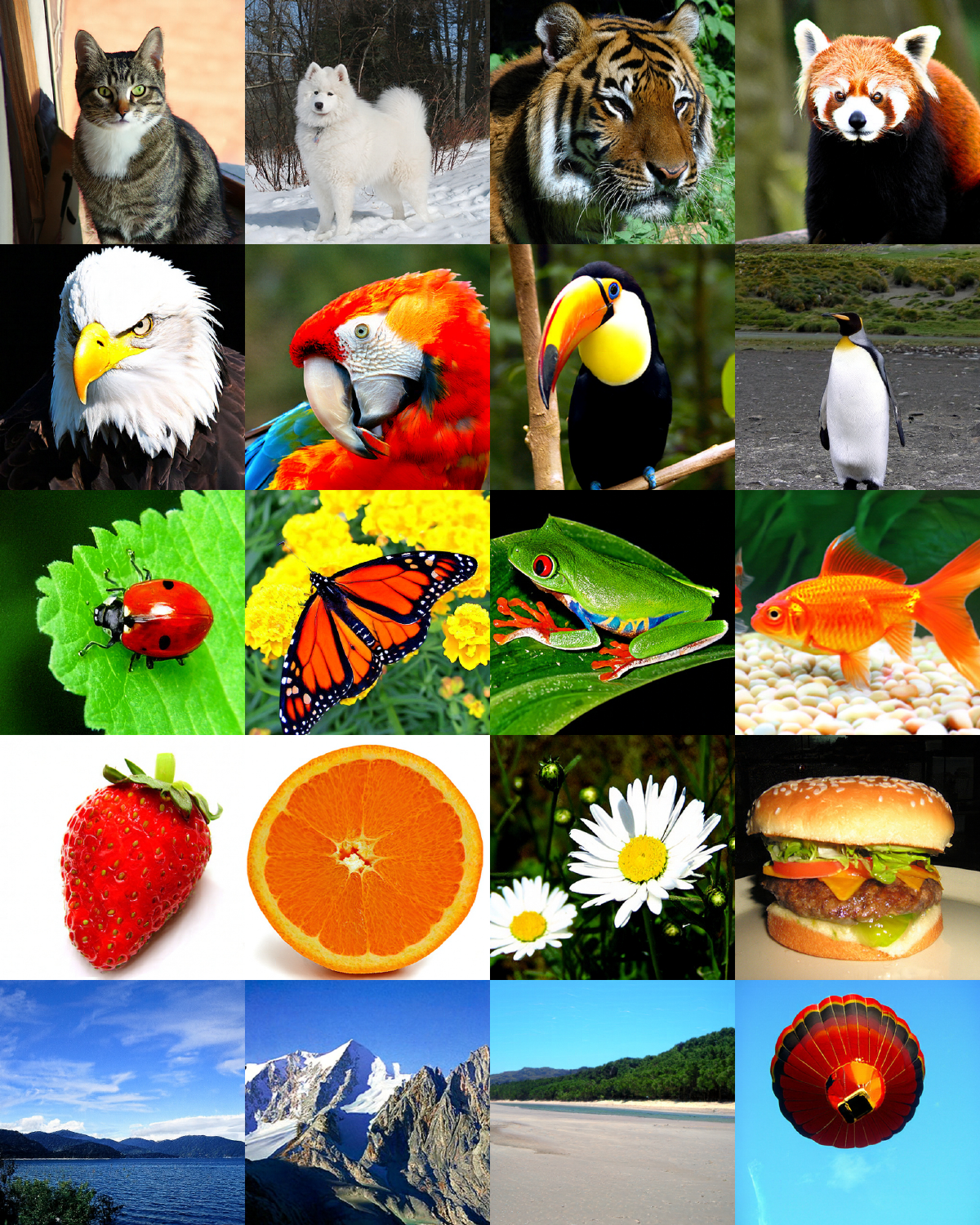}
    \caption{$512\times512$ samples of PixelDiT-XL/16+WiD at 260\,+\,50 epochs. We use time-shift 2.0 and classifier-free guidance
 with~$w = 4.0$.}
    \label{fig:512_samples}
\end{figure}

\begin{figure}[H]
    \centering
    \includegraphics[width=1\linewidth]{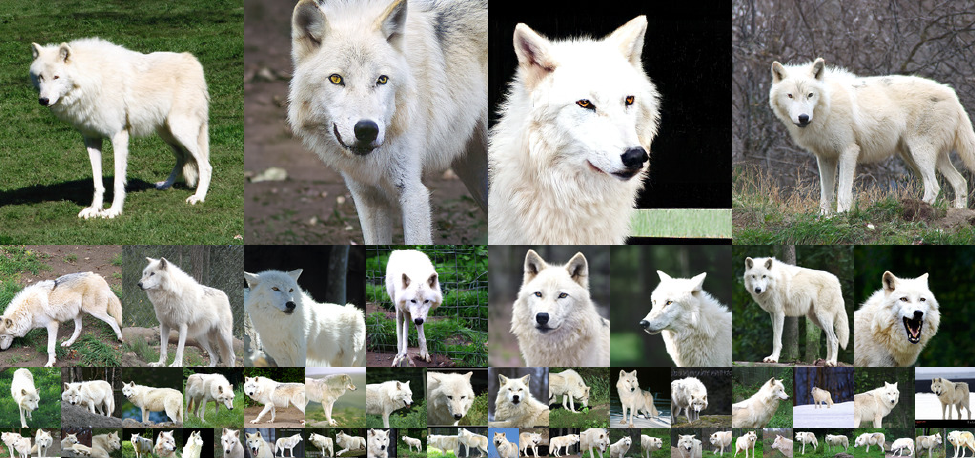}
    \caption{$256\times256$ samples of PixelDiT-XL/16+WiD at 260 epochs. We use classifier-free guidance
 with~$w = 4.0$.  Class label = ``arctic wolf'' (270).}
    \label{fig:vis270}
\end{figure}

\vspace{90pt}
\begin{figure}[H]
    \centering
    \includegraphics[width=1\linewidth]{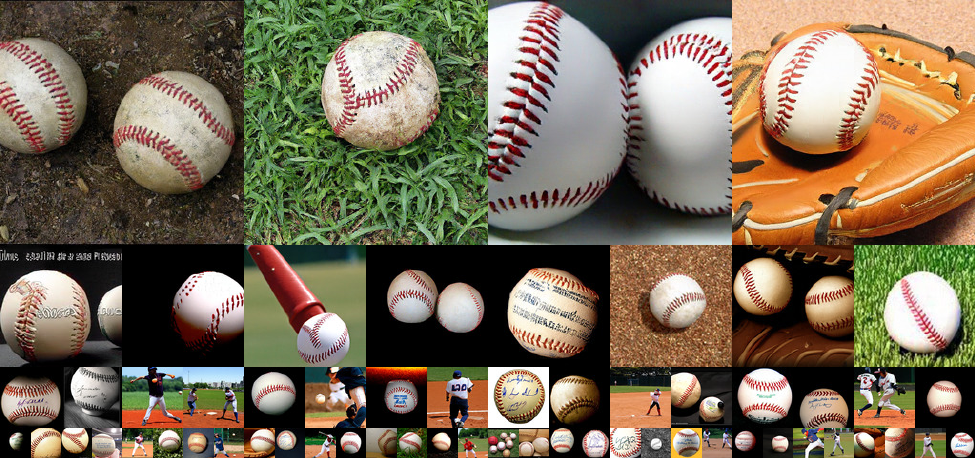}
    \caption{$256\times256$ samples of PixelDiT-XL/16+WiD at 260 epochs. We use classifier-free guidance
 with~$w = 4.0$. Class label = ``baseball'' (429).}
    \label{fig:vis429}
\end{figure}

\clearpage
\begin{figure}[p]
    \centering
    \includegraphics[width=1\linewidth]{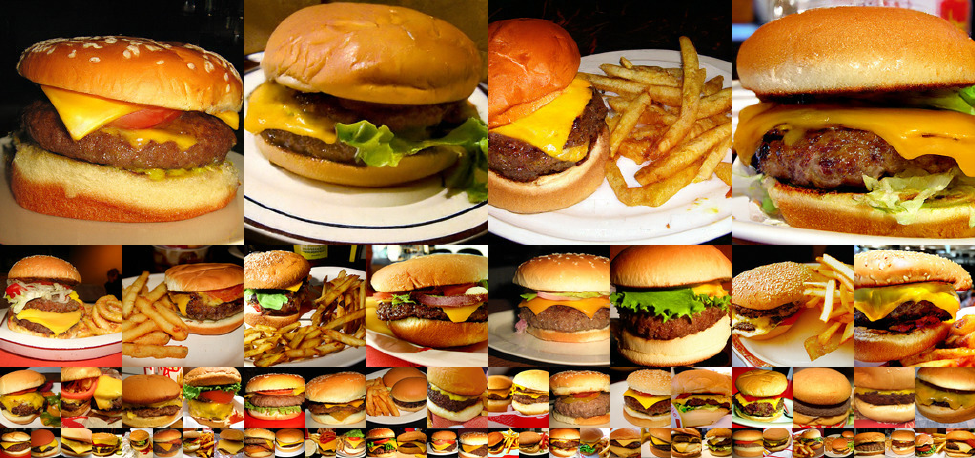}
    \caption{$256\times256$ samples of PixelDiT-XL/16+WiD at 260 epochs. We use classifier-free guidance
 with~$w = 4.0$.  Class label = ``cheeseburger'' (933).}
    \label{fig:vis933}
\end{figure}

\begin{figure}[p]
    \centering
    \includegraphics[width=1\linewidth]{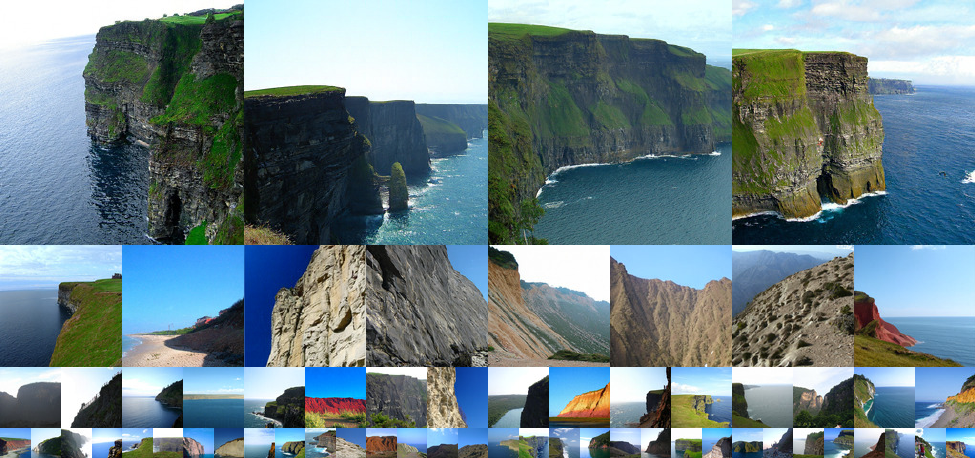}
    \caption{$256\times256$ samples of PixelDiT-XL/16+WiD at 260 epochs. We use classifier-free guidance
 with~$w = 4.0$. Class label = ``cliff drop-off'' (972).}
    \label{fig:vis972}
\end{figure}

\end{document}